\RequirePackage[table]{xcolor}
\documentclass[journal,twoside,web]{ieeecolor}

\usepackage{generic}
\usepackage{cite}
\usepackage{amsmath,amssymb,amsfonts}
\usepackage{algorithm,algorithmic}
\usepackage{graphicx}
\usepackage{textcomp}
\usepackage{multirow}
\usepackage{url}
\usepackage{booktabs, tabularx}
\usepackage{float}
\usepackage[hidelinks]{hyperref}

\definecolor{guidebluecell}{RGB}{204,229,255}
\definecolor{guidegreencell}{RGB}{220,237,200}
\definecolor{guidepeachcell}{RGB}{255,218,185}
\definecolor{guidepurplecell}{RGB}{211,208,253}
\definecolor{guideyellowcell}{RGB}{255,242,204}
\definecolor{guidemintcell}{RGB}{204,236,230}
\definecolor{guidepinkcell}{RGB}{248,218,232}

\def\BibTeX{{\rm B\kern-.05em{\sc i\kern-.025em b}\kern-.08em
    T\kern-.1667em\lower.7ex\hbox{E}\kern-.125emX}}
    
\begin{document}

\title{GUIDE: Reinforcement Learning for Behavioral Action Support in Type 1 Diabetes}
\author{Saman Khamesian\textsuperscript{1,2}*, Sri Harini Balaji\textsuperscript{3}, Di Yang Shi\textsuperscript{4}, Stephanie M. Carpenter\textsuperscript{1}, Daniel E. Rivera\textsuperscript{3}, W. Bradley Knox\textsuperscript{4}, Peter Stone\textsuperscript{4}, Hassan Ghasemzadeh\textsuperscript{1}
\thanks{$^{1}$College of Health Solutions, Arizona State University, Phoenix, USA}
\thanks{$^{2}$School of Computing and Augmented Intelligence, Arizona State University, Tempe, USA}
\thanks{$^{3}$School for Engineering of Matter Transport and Energy, Arizona State University, Tempe, USA}
\thanks{$^{4}$Department of Computer Science, The University of Texas at Austin, Austin, USA}
\thanks{{*\textbf{Corresponding author}: \textcolor{blue}{skhamesi@asu.edu}}}}

\maketitle

\begin{abstract}
Type 1 diabetes (T1D) management requires continuous adjustment of insulin and lifestyle behaviors to maintain blood glucose within a safe target range. Although automated insulin delivery (AID) systems have improved glycemic outcomes, many patients still fail to achieve recommended clinical targets. Current reinforcement learning (RL)-based methods focus primarily on insulin-only treatment and do not provide behavioral recommendations for glucose control. To address this gap, we propose $GUIDE$, an RL-based decision-support framework designed to complement AID technologies by providing structured behavioral recommendations defined by intervention type, magnitude, and timing, including bolus insulin administration and carbohydrate intake events. $GUIDE$ integrates a patient-specific glucose predictor trained on real-world continuous glucose monitoring data and supports offline and online RL algorithms within a unified environment. The algorithms are evaluated across $25$ individuals from the $\text{AZT1D}$ dataset and $12$ individuals from the $\text{OhioT1DM}$ dataset. Among the evaluated algorithms, $\text{CQL-BC}$ achieved the best performance on both datasets, with mean time-in-range values of $84.18 \pm 19.89\%$ on $\text{AZT1D}$ and $77.64 \pm 8.81\%$ on $\text{OhioT1DM}$. It also maintained time-below-range values of $0.43 \pm 1.27\%$ and $2.81 \pm 3.56\%$, respectively. Behavioral analysis yielded mean cosine similarities of $0.767 \pm 0.128$ on $\text{AZT1D}$ and $0.774 \pm 0.162$ on $\text{OhioT1DM}$, indicating that the learned policy preserves key structural characteristics of patient action patterns. These findings demonstrate the potential of conservative offline RL with a structured behavioral action space to provide personalized and behaviorally plausible decision support for diabetes management.
\end{abstract}


\section{Introduction}
\label{sec:introduction}

\IEEEPARstart{T}{ype} 1 Diabetes (T1D) is an autoimmune disease characterized by the destruction of pancreatic \(\beta-\)cells, resulting in insufficient endogenous insulin production and a lifelong requirement for exogenous insulin therapy~\cite{popoviciu2023type, o2024shifting}. It is estimated that 9.5 million people worldwide are living with T1D, including 1 million children aged 0–14 years and 0.8 million adolescents aged 15–19 years, representing a 13\% increase since 2021~\cite{ogle2025global}. Maintaining optimal glucose control is complex, given the daily variability in insulin needs driven by sleep patterns, meals, exercise, and other factors \cite{battelino2019clinical}. The primary objective of blood glucose level (BGL) control in T1D management is to maximize time spent within the target glycemic range ($70$--$180$ mg/dL) while minimizing exposure to hypoglycemia (BGL $<70$ mg/dL) and hyperglycemia (BGL $>180$ mg/dL), which correspond to clinically established glycemic thresholds~\cite{battelino2019clinical, foster2019state, american20226}.

\begin{figure}[t]
    \centering
    \includegraphics[width=0.8\linewidth]{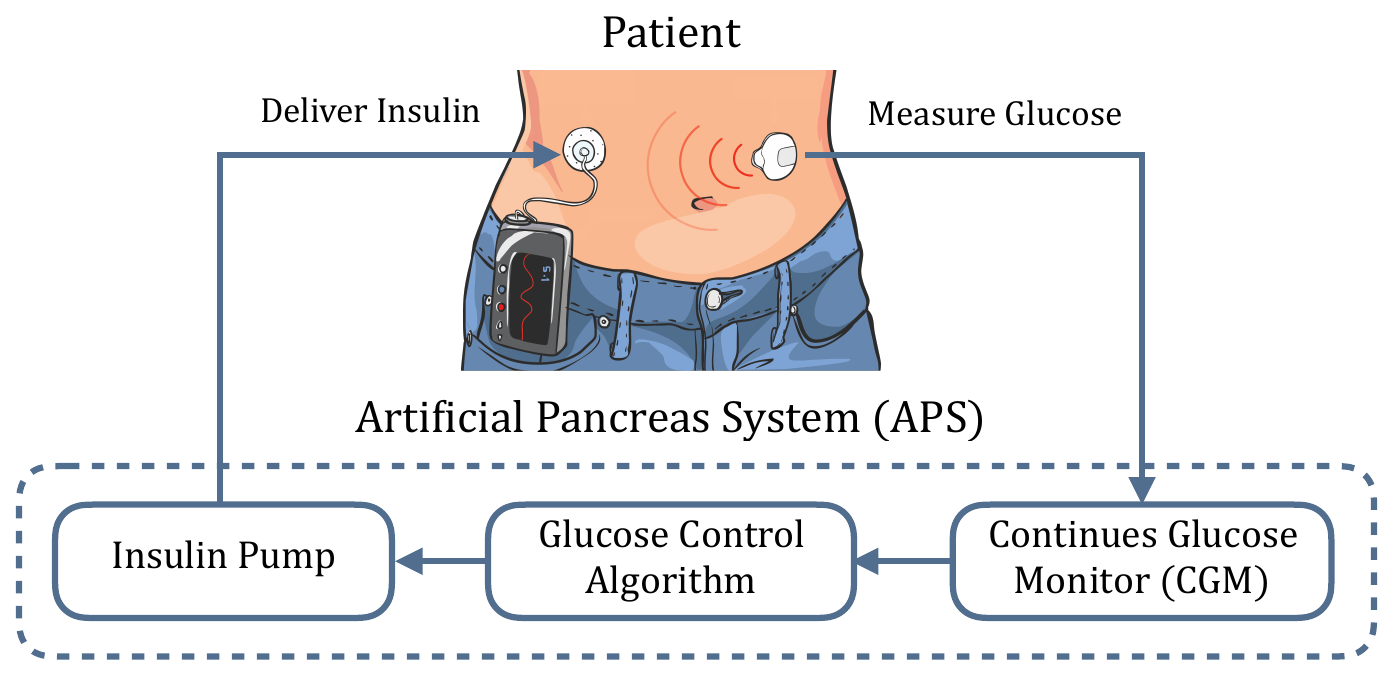}
    \caption{Illustration of an artificial pancreas system with closed-loop blood glucose control in patients with T1D. Glucose measurements from a continuous glucose monitor are processed by a control algorithm to compute insulin dosing delivered through an insulin pump.}
    \label{fig:artificial_pancreas}
\end{figure}

Control strategies for Artificial Pancreas System (APS) represent a significant milestone in diabetes management, aiming to automate insulin administration and reduce the burden on individuals with T1D~\cite{bekiari2018artificial}. Over the past decade, hybrid closed-loop (HCL) technologies have emerged as a major advancement toward this goal and are the current clinical realization of artificial pancreas systems~\cite{templer2022closed}. HCL systems integrate continuous glucose monitoring (CGM)~\cite{rodbard2016continuous} with automated insulin delivery (AID) technology~\cite{limbert2024automated}, and typically employ control strategies such as proportional-integral-derivative (PID)~\cite{steil2013algorithms} and model predictive control (MPC)~\cite{bequette2013algorithms} to regulate BGL. Fig.~\ref{fig:artificial_pancreas} illustrates an APS for T1D management.

Despite substantial technological progress, nearly 80\% of the overall T1D population does not meet key clinical targets recommended by the American Diabetes Association, such as achieving at least 70\% time-in-range (TIR) and maintaining an HbA1c value near 7\% \cite{battelino2019clinical, foster2019state, american20226}. Even among patients using AID systems, approximately 30\% do not achieve these targets \cite{bombaci2025impact}. To further improve the performance and adaptability of AID systems within APS, recent research has explored reinforcement learning (RL) as a learning-based approach for insulin control \cite{tejedor2020reinforcement, denes2024reinforcement, bothe2013use}. In existing RL-based approaches for insulin delivery, the agent typically observes recent glucose measurements and insulin delivery history and learns a policy that maps these observations to insulin dosing decisions, with the objective of maintaining glucose levels within the target range. By adapting dosing decisions to patient-specific dynamics and disturbances such as meals or changes in insulin sensitivity, RL provides a framework for learning personalized glucose control strategies. In simulated environments, RL-based controllers have demonstrated the ability to learn patient-specific policies and achieve strong performance compared with standard PID and MPC algorithms\cite{tejedor2020reinforcement, denes2024reinforcement, bothe2013use, aria2025learning}.

Recent RL-based solutions have sought to reduce patient burden by enabling agents to infer insulin needs directly from glucose data, thereby minimizing the need for explicit user input \cite{fox2020deep, hettiarachchi2024g2p2c, lee2020toward}. Nevertheless, real-world glycemic control remains strongly shaped by everyday self-management behaviors and lifestyle factors beyond insulin delivery alone, including physical activity, sleep patterns, and the temporal relationship between eating and insulin administration. Consequently, human factors such as delayed bolus delivery, inappropriate carbohydrate intake, or inaccurate carbohydrate estimation frequently lead to episodes of hyperglycemia and hypoglycemia that cannot be fully eliminated by algorithmic solutions alone. Achieving effective T1D management therefore benefits not only from AID systems but also from intelligent decision tools that assist patients in self-management decisions around bolus insulin administration and snack intake to minimize risks of hyperglycemia and hypoglycemia.

\begin{figure}[t]
    \centering
    \includegraphics[width=0.8\linewidth]{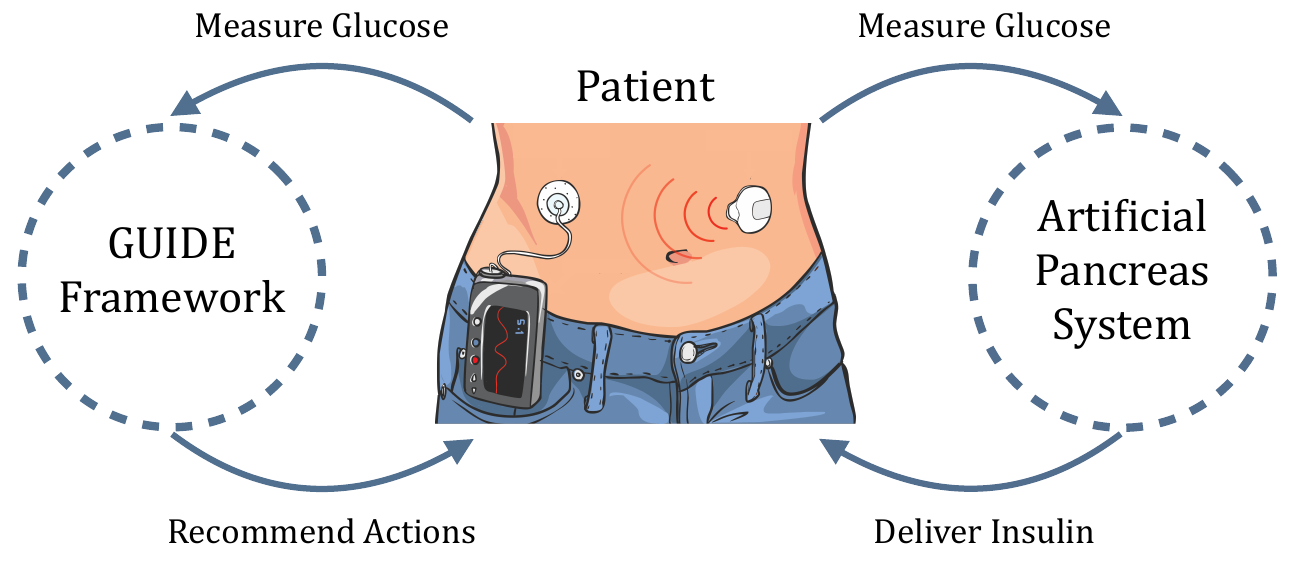}
    \caption{Schematic overview illustrating the complementary roles of the AID system and the GUIDE framework, with both components utilizing glucose measurements to inform insulin delivery and behavioral action recommendations for patients with T1D.}
    \label{fig:concept}
\end{figure}

To address this gap, we present GUIDE: \textbf{G}l\textbf{u}cose \textbf{I}ntelligence \textbf{D}ecision \textbf{E}ngine, an RL-based framework that generates patient-specific recommendations for insulin and carbohydrate intake, adapting to individual glucose dynamics and daily routines (Fig.~\ref{fig:concept}). GUIDE supports a behavioral action space in which the decision policy operates over multiple self-management behaviors rather than insulin adjustment alone. Each action specifies the behavioral type (e.g., bolus insulin or carbohydrate intake), along with its timing and magnitude, enabling structured and patient-specific recommendations. In addition, a bolus insulin safeguard filters each proposed dose before execution to limit insulin stacking and excessive cumulative insulin delivery. Finally, we devise different RL algorithms within the proposed GUIDE framework and evaluate their performance through a personalized data-driven simulation environment. This environment integrates a glucose-level predictor trained on each patient’s data, a glucose-responsive basal insulin generator, and a human-inspired meal generator to simulate individualized glucose dynamics and realistic meal patterns. This controlled evaluation enables direct comparison of decision-making policies while avoiding the safety risks, ethical constraints, high costs, and time-consuming testing of new strategies in real patients.

\section{Related Work}
\label{sec:related_Work}

The application of RL in healthcare has been growing steadily, showing significant results in the management of T1D management \cite{tejedor2020reinforcement, denes2024reinforcement, bothe2013use}. Traditional insulin delivery strategies, such as PID and MPC controllers, have provided an important foundation for automated insulin delivery systems. These model-based control approaches typically rely on system identification or manual tuning to adapt to patient-specific dynamics \cite{shah2016insulin, doyle2014closed, ramkissoon2018unannounced}.

To overcome these challenges, a growing body of work has explored advanced RL-based approaches. For instance, Emerson et al. \cite{emerson2023offline} introduced an offline RL framework for safer blood glucose control in people with T1D, evaluating algorithms such as Twin Delayed Deep Deterministic Policy Gradient with Behavior Cloning (TD3-BC) \cite{fujimoto2018addressing}, Batch-Constrained Q-Learning (BCQ) \cite{fujimoto2019off}, and Conservative Q-Learning (CQL) \cite{kumar2020conservative}. Their study showed that offline RL can learn effective dosing policies directly from retrospective data, improving TIR and patient safety without the risks associated with online exploration. Importantly, their approach proved robust to common real-world challenges, such as irregular meal schedules and sensor errors, highlighting the potential of offline RL for real clinical applications.

Adjevi et al.~\cite{adjevi2026explainable} proposed an explainable glucose-monitoring framework that combines a CNN-based predictor with SHAP-guided Actor--Critic reinforcement learning. SHAP values quantify the contributions of multimodal input features to the predicted glucose level and identify the feature to be adjusted, while the RL component applies bounded modifications to the selected feature to move the predicted glucose toward a target range. The framework was evaluated as a research-only simulation, and the authors emphasized that the SHAP attributions represent associative feature contributions rather than causal clinical relationships.

Recent efforts have also introduced modular RL controllers designed to further personalize and safeguard automated insulin delivery. Marchetti et al. \cite{marchetti2025deep} proposed an RL system based on a Dual Proximal Policy Optimization (Dual-PPO) architecture, where two separate agents are responsible for hyperglycemic and euglycemic ranges, supported by a built-in safety mechanism to suspend insulin during hypoglycemic episodes. While this dual-agent setup adapts dosing to different glycemic conditions and outperforms single-agent RL and classical methods, it still relies on meal announcements, similar to other hybrid closed-loop approaches.

\begin{figure*}[!htbp]
\centering
\includegraphics[width=0.8\linewidth]{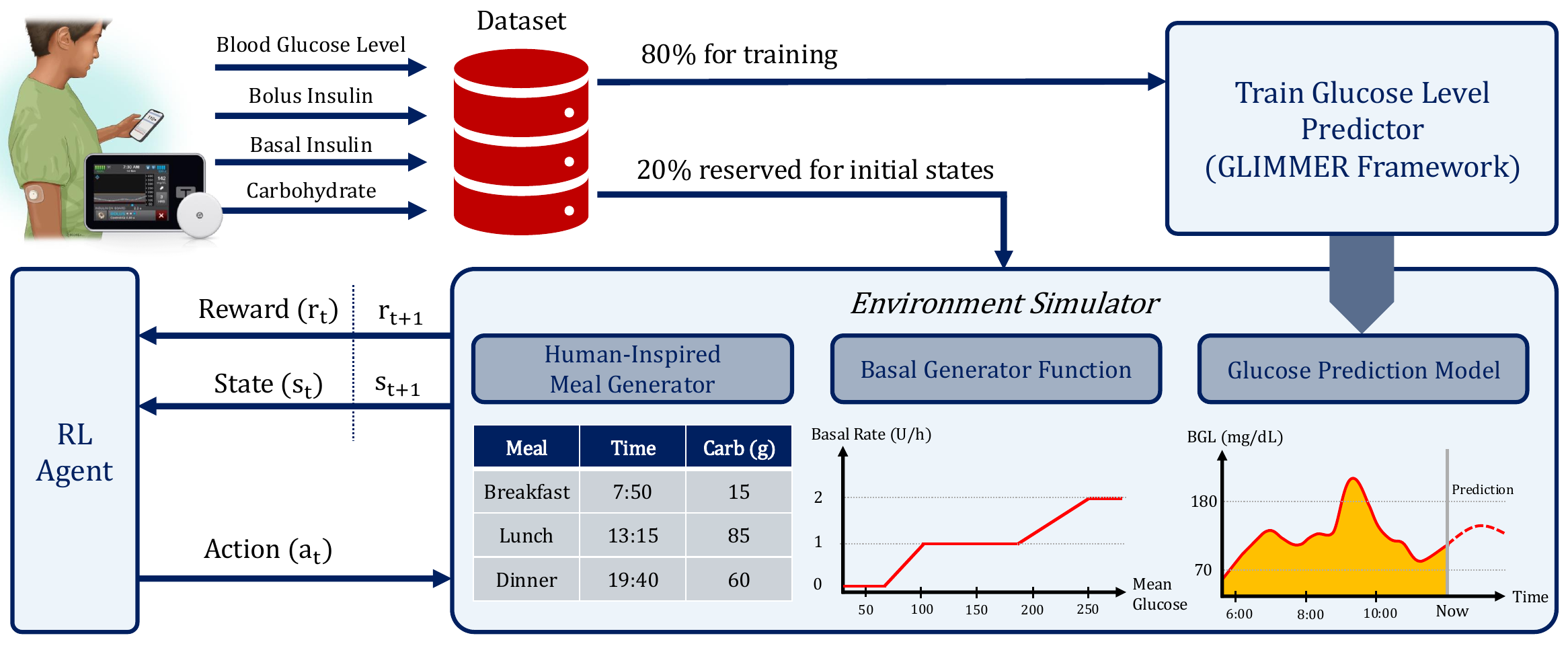}
\caption{Overview of the GUIDE framework. Data are partitioned chronologically within each subject, with 80\% used to train the glucose level predictor (GLIMMER) and 20\% reserved for defining initial states. The trained predictor is deployed within the environment simulator alongside the human-inspired meal generator and glucose-responsive basal insulin generator. At each decision step, the RL agent selects a behavioral action based on the current state, and the environment returns the resulting reward and next state, forming a closed-loop learning process.}
\label{fig:framework}
\vspace{-8pt}
\end{figure*}

Another major direction has focused on eliminating manual meal announcements, a longstanding burden for users of diabetes technology. Hettiarachchi et al. \cite{hettiarachchi2024g2p2c} introduced G2P2C, a modular RL framework that combines a glucose prediction module with a planning controller, allowing the system to anticipate glucose excursions and adjust insulin dosing without requiring the user to announce meals. G2P2C leverages recent CGM and insulin data to infer potential meal events and generate personalized dosing strategies in real time, thereby improving postprandial glucose control. Bioinspired RL \cite{lee2020toward} approaches further advance automation by shaping the agent’s reward structure to reflect physiological homeostasis, often incorporating knowledge of insulin pharmacokinetics and the body’s adaptive glucose regulation. In parallel, deep RL frameworks \cite{fox2020deep} have integrated temporal context and pattern recognition, enabling the agent to learn and predict glucose dynamics across varying scenarios—such as unannounced meals—without manual intervention.

Many existing RL-based glucose-control methods are developed and evaluated using the UVA/Padova simulator~\cite{man2014uva}, which provides a standardized environment based on synthetic virtual subjects and enables controlled comparison before real-world evaluation. Although this is valuable for algorithm development, synthetic cohorts may not fully reflect the variability of real-world patient trajectories and behaviors. In contrast, GUIDE uses personalized simulation environments built around glucose prediction models trained on real-world patient data, allowing initialization from observed trajectories and incorporation of behavioral signals.

\section{Methodology}
\label{sec:methodology}

We develop GUIDE as an RL-based framework for closed-loop blood glucose control in individuals with T1D. The framework consists of an environment simulator and an RL agent operating over a behavioral action space (Fig.~\ref{fig:framework}). The environment incorporates a personalized glucose level predictor trained on real-world patient data to model individualized glucose dynamics, a human-inspired meal generator that simulates realistic daily meal patterns, and a glucose-responsive basal insulin generator that simulates basal delivery throughout each rollout. The RL agent interacts with this environment to learn decision policies that recommend a single behavioral action at each decision step.

We assume that the RL agent issues recommendations at 1-hour intervals, even though behavioral and physiological data may be sampled at higher temporal resolutions (e.g., CGM provides measurements at 5-minute intervals). This design choice is motivated by both physiological and practical considerations. First, GUIDE is designed to prevent hyperglycemic and hypoglycemic events, which typically occur on the order of hours rather than minutes. Consequently, finer intervention granularity is unlikely to yield meaningful additional benefit. Second, the effects of behavioral and therapeutic actions—such as food intake and insulin administration—are inherently delayed and temporally diffuse. For example, postprandial glucose responses typically begin 15–30 minutes after a meal and evolve over a longer time horizon. Due to this delayed system response, issuing recommendations at overly short intervals may lead to redundant or conflicting actions, potentially reducing patient adherence and increasing the risk of overtreatment.

We further assume full patient compliance with recommended actions in the simulation environment used for policy training. This assumption allows us to focus on the core design of the GUIDE framework without introducing additional uncertainty related to human behavior during training. Accordingly, the action space focuses on modeling the type, magnitude, and timing of recommended interventions, without explicitly representing deviations from recommendations (e.g., ignoring suggestions or delaying actions). Modeling adherence to behavioral interventions involves complex, context-dependent factors—including motivation, cognition, and environmental constraints—and represents a substantial research challenge in its own right. While incorporating realistic models of patient compliance into the simulator and policy-training process is an important direction for future work, it is beyond the scope of the present study, which focuses on the algorithmic and systems-level aspects of the proposed approach.

\subsection{Problem Statement}
\label{sec:problem_statement}

To formally describe the decision-making process in GUIDE, we formulate the glycemic management problem as a Markov Decision Process (MDP) defined by the tuple
\(
\mathcal{M} = (\mathcal{S}, \mathcal{A}, \mathcal{P}, \mathcal{R}, \gamma),
\) where:
\begin{itemize}
    \item \( \mathcal{S} \) is the state space,
    \item \( \mathcal{A} \) is the action space,
    \item \( \mathcal{P}: \mathcal{S} \times \mathcal{A} \rightarrow \mathcal{S} \) denotes the transition dynamics,
    \item \( \mathcal{R}: \mathcal{S} \times \mathcal{A} \rightarrow \mathbb{R} \) is the reward function,
    \item \( \gamma \in (0, 1] \) is the discount factor.
\end{itemize}
In practice, the dynamics of glucose metabolism are only partially observable: the agent cannot access the true physiological state of the patient, but instead receives proxy measurements such as CGM signals, meal records, and insulin doses. This makes the problem closer to a Partially Observable Markov Decision Process (POMDP). To mitigate this limitation, we approximate the ground truth state by aggregating temporal histories, contextual features, and elapsed times since past interventions, forming \( s_t \in \mathcal{S} \). At each decision step \( t \), the agent observes a state \( s_t \), selects an action \( a_t \in \mathcal{A} \), receives a reward \( r_t \in \mathbb{R} \), and transitions to the next state \( s_{t+1} \), forming tuples 
\((s_t, a_t, r_t, s_{t+1}, d_t)\), where \( d_t \in \{0,1\} \) indicates episode termination. The goal is to learn a deterministic policy \(\pi: \mathcal{S} \rightarrow \mathcal{A} \) that maximizes the expected discounted return commonly defined as:
\[
\mathbb{E} \left[ \sum_{t=0}^{T} \gamma^t r_t \right].
\]

\subsection{State Space}
\label{sec:state_space}

The state at decision time \(t\) is represented as a concatenation of eight feature vectors, each containing values from the past six hours at 5-minute resolution:
\[
s_t = \big[H,\ Z,\ G,\ C,\ I^{\text{bolus}},\ I^{\text{basal}},\ \Delta^{\text{meal}},\ \Delta^{\text{inject}}\big],
\]
where:
\begin{itemize}
    \item \(H\) denotes the hour-of-day, with \(h_t \in \{0,\ldots,23\}\),
    \item \(Z\) denotes the binary sleep indicators, with \(z_t \in \{0,1\}\),
    \item \(G\) denotes glucose values (mg/dL),
    \item \(C\) denotes carbohydrate intake (grams),
    \item \(I^{\text{bolus}}\) denotes bolus insulin (units),
    \item \(I^{\text{basal}}\) denotes basal insulin (U/h),
    \item \(\Delta^{\text{meal}}\) denotes the elapsed times since the last meal,
    \item \(\Delta^{\text{inject}}\) denotes the elapsed times since the last bolus insulin injection.
\end{itemize}

In addition to glucose values, bolus and basal insulin, and carbohydrate intake, we incorporate the temporal features \(\Delta^{\text{meal}}\) and \(\Delta^{\text{inject}}\) to capture the delay and timing of recent meal and bolus insulin actions. Also, the contextual variables \(H\) and \(Z\) provide circadian information, enabling the agent to differentiate daytime and nighttime conditions and typical daily timing patterns.

\subsection{Action Space}
\label{sec:action_space}

Each action \(a_t \in \mathcal{A}\) is represented as a six-dimensional vector:
\[
a_t = [s_{\text{nothing}},\ s_{\text{eat}},\ s_{\text{inject}},\ c_t,\ i_t,\ k_t].
\]
The design reflects both high-level action categories and fine-grained control:

\begin{itemize}
    \item The first three elements are real-valued scores representing the agent's preference among three action types: do nothing, eat, or inject insulin. The final decision is obtained by applying \textit{argmax} to these scores during inference.
    \item \(c_t \in [c_{\min},c_{\max}]\) denotes the amount of carbohydrate to consume if a meal is selected.
    \item \(i_t \in [i_{\min},i_{\max}]\) denotes the bolus insulin amount if an injection is selected.
    \item \(k_t \in \{0,\ldots,11\}\) denotes the five-minute index within the next hour when the action is applied.
\end{itemize}

Compared to many prior RL-based glucose-control studies that restrict actions to simplified abstractions (e.g., fixed insulin adjustments or threshold-based rules), this formulation defines a richer representation that jointly models action type, magnitude, and timing. The numerical bounds can be selected based on empirical distributions observed in historical data from the target patient cohort.

\subsection{Bolus Insulin Safeguard}
\label{sec:bolus_safeguard}

The bolus range defined in the action space represents the range within which the RL agent can propose an insulin action. To prevent violations of bolus insulin constraints and strengthen insulin-delivery safety, we introduce a safeguard layer that is applied to each proposed dose before execution in the simulation environment. The safeguard combines estimated insulin-on-board, a rolling cumulative bolus limit, and prevention of insulin stacking through a minimum bolus interval. This design is consistent with prior simulation-based glucose-control studies that incorporate insulin activity or explicit insulin-safety mechanisms into the decision process~\cite{emerson2023offline,marchetti2025deep}.

Let \(\hat{i}_t\) denote the bolus proposed by the policy. The maximum allowable bolus at decision time \(t\) is defined as
\begin{equation}
U_t
=
S_t
\min
\left\{
i_{\max},
\left[I_{\max}-\mathrm{IOB}_t\right]^+,
\left[B_{\max}-B_t(W)\right]^+
\right\},
\end{equation}
where \(\mathrm{IOB}_t\) denotes estimated insulin-on-board, \(B_t(W)\) denotes cumulative bolus insulin over the rolling window \(W\), \(S_t\) is the stacking indicator, \(i_{\max}\) is the upper bound of the bolus action space, and \(I_{\max}\) and \(B_{\max}\) are the active-insulin and cumulative-bolus limits, respectively. Here, \([x]^+=\max(0,x)\), and \(S_t\) equals zero when the minimum bolus interval is violated and one otherwise. The safeguarded bolus is then obtained as
\begin{equation}
z_t=\operatorname{clip}\left(\hat{i}_t,0,U_t\right).
\end{equation}

If \(U_t\) falls below the minimum bolus amount \(i_{\min}\), the injection is suppressed and the \textit{Nothing} action is executed, preserving the original action space. Otherwise, \(z_t\) is executed in the simulation environment. This safeguard is designed specifically for the GUIDE simulation environment and is not presented as a patient-specific or clinically validated insulin-dosing mechanism.

\subsection{Reward Function}
\label{sec:reward_function}

The reward function in RL must align with the clinical objectives of the task. In the context of T1D, reward design is often framework-specific and lacks standardization, reflecting the absence of a universally accepted formulation across prior studies~\cite{tejedor2020reinforcement}. Accordingly, we define a reward function that integrates glycemic regulation with clinically motivated behavioral considerations. Rewards are computed at an hourly resolution within a 24-hour episode. At each decision time $t$, the reward is defined as:

\begin{equation}
\begin{aligned}
r_t
&=
r_{\text{glucose}}(t)
+ r_{\text{meal}}(s_t, a_t)
+ r_{\text{insulin}}(s_t, a_t) \;+ \\
&\quad
\;r_{\text{sleep}}(s_t, a_t)
+ r_{\text{stability}}(s_t, a_t),
\end{aligned}
\end{equation}
where $s_t$ represents the system state and $a_t$ denotes the action selected by the agent.

The glycemic component is defined using clinically established thresholds for hypoglycemia and hyperglycemia~\cite{battelino2019clinical, foster2019state, american20226}, with asymmetric slopes informed by prior glucose-specific loss design~\cite{khamesian2026glycemic}:

\begin{equation}
\tilde{r}_\text{g}(g) =
\begin{cases}
-7(70-g), & g < 70, \\[6pt]
100\left(1 - \dfrac{|g-125|}{55}\right), & 70 \le g \le 180, \\[6pt]
-2(g-180), & g > 180.
\end{cases}
\end{equation}

At each decision time $t$, the glycemic reward is computed by summing the rewards evaluated at the predicted glucose values over the next hour (12 five-minute intervals):

\begin{equation}
r_{\text{glucose}}(t)
=
\sum_{k=1}^{12}
\tilde{r}_{\text{g}}\!\left(g_{t+k}\right).
\end{equation}

The remaining reward components encode clinically motivated behavioral considerations related to timing, contextual safety, and behavioral stability, and share a unified structure:

\begingroup
\small
\begin{equation}
r_k(s_t, a_t)
=
\sum_{j \in \mathcal{J}_k}
\psi_j(s_t, a_t),\
k \in
\{\text{meal},\,\text{insulin},\,\text{sleep},\,\text{stability}\}.
\end{equation}
\endgroup

Each index set $\mathcal{J}_k$ defines a group of clinically motivated rules associated with component $k$. For action-dependent components, the rules are conditionally activated based on the selected action (e.g., $a_t = \texttt{EAT}$ or $a_t = \texttt{INJECT}$). Specifically, $\mathcal{J}_{\text{meal}}$ includes glucose- and timing-dependent effects associated with carbohydrate intake; $\mathcal{J}_{\text{insulin}}$ captures glucose level, glucose trend, and recent meal or insulin context for bolus administration; $\mathcal{J}_{\text{sleep}}$ encodes behavior during sleep mode; and $\mathcal{J}_{\text{stability}}$ captures penalties for omitting a needed action, rewards stable glucose without intervention, and discourages repeated non-$\texttt{NOTHING}$ actions.

Individual rule terms $\psi_j$ are expressed as linear or piecewise-linear functions of clinically relevant variables, with clipping applied where specified:

\begin{equation}
\psi_j(s_t, a_t)
=
\mathrm{clip}
\big(
\alpha_j \cdot f_j(s_t, a_t),
\, L_j,
\, U_j
\big),
\end{equation}
where
\begin{itemize}
    \item $f_j(\cdot)$: linear or piecewise-linear function of clinically relevant variables (e.g., glucose deviation, time since the previous meal or insulin administration);
    \item $\alpha_j$: scaling coefficient controlling the strength of the response;
    \item $L_j, U_j$: lower and upper clipping limits applied to the rule term.
\end{itemize}

Each rule is activated only when its corresponding condition is satisfied. For example, in the meal-related component, a rule may penalize actions taken shortly after a previous meal by defining $f_j$ as the elapsed time since the last meal and applying a penalty when this interval falls below a predefined threshold.

\vspace{-5pt}
\subsection{Environment Simulator}
\label{sec:env_simulator}

The environment simulator generates the next state at each time step based on the current state and applied actions. It consists of three components: (i) a glucose-responsive basal insulin generator that simulates basal delivery throughout the rollout, (ii) a human-inspired meal generator that produces daily main meals based on predefined time windows and a truncated normal distribution, and (iii) a patient-specific glucose level predictor that forecasts glucose values.

At each step, the glucose level predictor generates the glucose values for the upcoming hour using the current state as input. The basal insulin generator determines the corresponding hourly basal rate, while time-dependent features such as time of day, sleep indicator, time since last meal, and time since last insulin are updated according to the applied action and simulation clock. The resulting glucose values, basal insulin rates, and updated contextual features are then used to construct the next state.

\subsubsection{Basal Insulin Generator}
\label{sec:basal_generator}

In real-world T1D management, basal insulin is delivered and adjusted by an insulin pump or AID system. Because no AID system operates within the data-driven simulation environment, we developed a glucose-responsive heuristic generator to reproduce this essential component of insulin delivery during the simulated 24-hour trajectory. The initial state contains the patient's recorded basal values, whereas basal rates during the rollout are generated dynamically according to the evolving glucose trajectory rather than remaining tied to historical values.

\begin{figure}[b!]
\centering
\includegraphics[width=0.85\columnwidth]{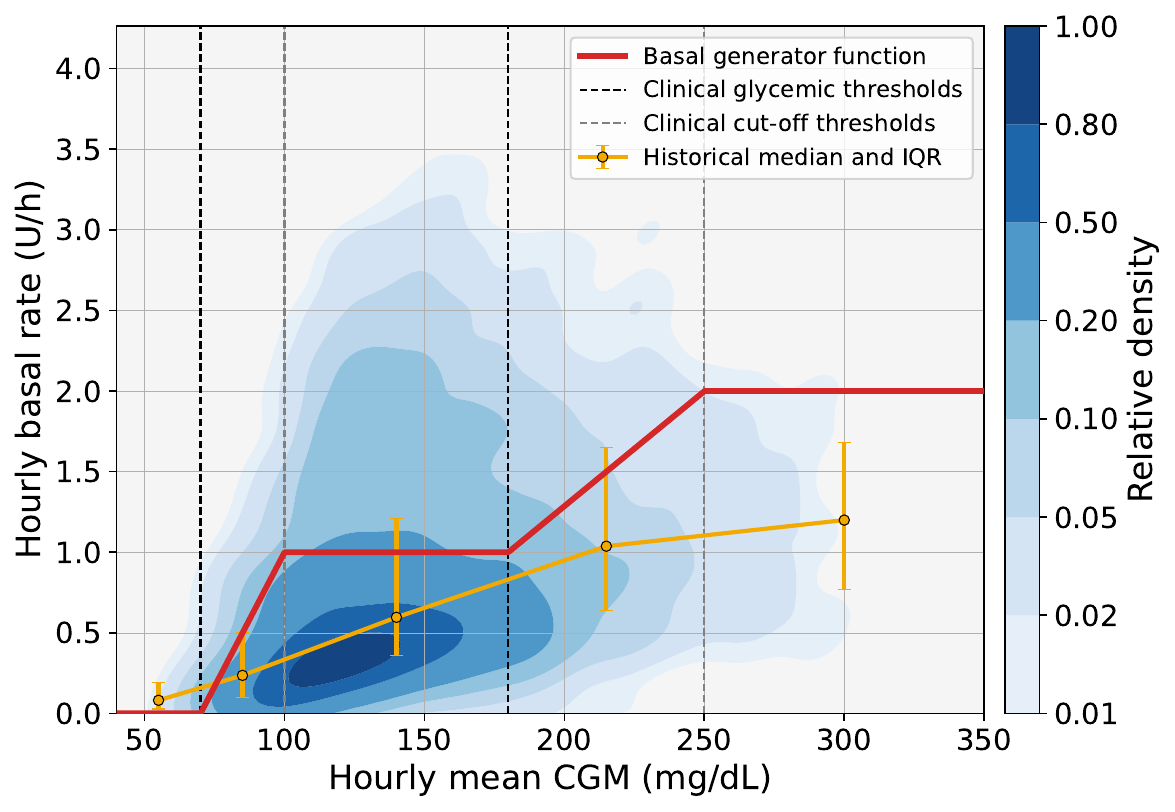}
\caption{Historical basal-rate distribution and basal generator function. Blue contours show the relative density of hourly mean CGM--basal-rate pairs, orange markers and error bars show the median and interquartile range, and the red line represents the generator function. Black and gray dashed lines mark the clinical glycemic thresholds (70 and 180 mg/dL) and clinical cut-off thresholds (100 and 250 mg/dL), respectively.}
\label{fig:basal_insulin_generator}
\end{figure}

At each hourly simulation step, the generator calculates the mean glucose over the upcoming hour and determines the basal rate using the piecewise heuristic function shown in Fig.~\ref{fig:basal_insulin_generator}. The function incorporates the clinically established hypoglycemia and hyperglycemia thresholds at 70 and 180 mg/dL and the clinical cut-off thresholds at 100 and 250 mg/dL \cite{battelino2019clinical}. Together, these thresholds define the gradual transitions in basal delivery as the mean glucose increases. The selected operating range was assessed against the historical training distribution, which exhibited the same overall increasing trend and contained the nominal 1 U/h rate and 2 U/h cap within the observed basal-rate range. These rates are bounded simulation parameters and are not intended to represent universal or patient-specific clinical dosing recommendations. Representing basal insulin as a dynamically generated component is also consistent with prior simulator-based glucose-management studies \cite{yamagata2020model,fox2020deep}.

\subsubsection{Human-Inspired Meal Generator}
\label{sec:meal_generator}

To simulate realistic main meal patterns as part of regular eating behavior, we introduce a human behavior controller that generates main meals (breakfast, lunch, and dinner) independently of the RL agent. At the start of each episode, the controller samples a meal hour for each meal type from plausible daily time windows (e.g., 7--9 a.m. for breakfast). When the simulation clock nears one of these scheduled hours, a meal is randomly assigned to a five-minute slot within the upcoming hour to avoid rigid timing. Carbohydrate amounts are drawn from a truncated normal distribution to reflect inter- and intra-individual variability while ensuring plausible values. The exact time windows and carbohydrate intake parameters are summarized in Table~\ref{tab:main_meal_protocol}. Once scheduled, the controller emits an \textit{EAT} action specifying the selected portion and time slot.

\begin{table}[htbp]
\centering
\footnotesize
\caption{Meal occurrence windows and CHO (carbohydrate) intake parameters used by the human-inspired meal generator. For each meal, a start hour is randomly selected from the specified time range. Carbohydrate portions are sampled from a truncated normal distribution (mean = 65 g, SD = 15 g, range = 20–100 g).}
\begin{tabularx}{\columnwidth}{Xcc}
\toprule
Meal & Time Window & CHO Distribution (g) \\
\midrule
Breakfast & 07:00–09:00 & \(\mathcal{TN}(65, 15^2,\ 20,\ 100)\) \\
Lunch     & 12:00–14:00 & \(\mathcal{TN}(65, 15^2,\ 20,\ 100)\) \\
Dinner    & 19:00–22:00 & \(\mathcal{TN}(65, 15^2,\ 20,\ 100)\) \\
\bottomrule
\end{tabularx}
\label{tab:main_meal_protocol}
\end{table}

\subsubsection{Glucose Prediction Model}
\label{sec:glimmer}

We incorporate a personalized glucose prediction model, denoted as GLIMMER (Glucose Level Indicator Model with Modified Error Rate), to forecast glucose levels \cite{khamesian2026glycemic}. GLIMMER is an architecture-agnostic forecasting framework with a custom loss designed to improve accuracy in dysglycemic regions, trained independently for each patient using real-world CGM, insulin, and carbohydrate intake records. Within the RL environment, GLIMMER is adapted to predict future glucose values based on the current state and selected action, and these predictions are used to construct the next state and simulate glucose dynamics during training and evaluation. The model outputs a 12-dimensional sequence of predicted glucose values for the next hour, with one value every 5 minutes:
\begin{equation}
Y = f_{\text{GLIMMER}}(X),
\end{equation}
where \(X = \langle x_{t-71}, \ldots, x_t \rangle\), with each \(x_k \in \mathbb{R}^6\) representing the feature vector at time \(k\):
\begin{itemize}
    \item \(g_k\): continuous glucose value (mg/dL),
    \item \(c_k\): carbohydrate intake (grams),
    \item \(i^{\text{bolus}}_k\): bolus insulin dose (units),
    \item \(i^{\text{basal}}_k\): basal insulin rate (U/h),
    \item \(g^{\text{cat}}_k \in \{\text{hypo},\ \text{normal},\ \text{hyper}\}\): glucose value classes,
    \item \(g^{\text{MA200}}_k\): Moving average of past 200 glucose values.
\end{itemize}
and \(Y = \langle y_{t+1}, \ldots, y_{t+12} \rangle\), where each \(y_j\) is the predicted glucose value at 5-minute intervals over the next hour.

\subsection{Reinforcement Learning Algorithms}
\label{sec:rl_algotihms}

The GUIDE framework evaluates reinforcement learning algorithms under two data-access regimes: offline training and online training \cite{alsadat2025trajectory}. TD3-BC~\cite{fujimoto2018addressing}, CQL-BC~\cite{kumar2020conservative}, and SAC-Offline are trained exclusively from a fixed transition dataset generated before policy optimization, whereas PPO~\cite{schulman2017proximal} and SAC-Online~\cite{haarnoja2018soft} collect new simulator transitions as training progresses. In this study, the term \textit{offline} refers to the absence of additional environment interaction during policy optimization; it does not indicate that the replay buffer consists of direct historical patient transitions. All algorithms use the same state and action representations, environment dynamics, and reward function, while differing in their policy optimization objectives and training procedures.

All evaluated algorithms follow an actor--critic structure, where a policy network $\pi_\theta(a \mid s)$ selects actions and a critic estimates the expected return. For the off-policy algorithms, the critic estimates the action-value function $Q_\phi(s,a)$ and is trained by minimizing a Bellman residual of the form
\begin{equation}
y_t = r_t + \gamma (1 - d_t)\,
\mathbb{E}_{a' \sim \pi(\cdot \mid s_{t+1})}
\left[Q'(s_{t+1},a')\right],
\end{equation}
with the corresponding loss
\begin{equation}
\mathcal{L}_{\text{critic}}
=
\mathbb{E}_{(s_t,a_t,r_t,s_{t+1})}
\left[
\left(Q(s_t,a_t)-y_t\right)^2
\right],
\end{equation}
where $\gamma$ is the discount factor and $d_t$ indicates episode termination. Target networks are used by the off-policy algorithms to stabilize training.

\subsubsection{Online Algorithms}
\label{sec:online_algo}

In the online setting, PPO and SAC-Online update their policies using transitions collected through ongoing interaction with the simulator. PPO follows an on-policy training procedure in which newly collected trajectories are used for policy updates. It employs a clipped surrogate objective to promote stable policy improvement:
\begin{equation}
\mathcal{L}_{\text{PPO}}
=
\mathbb{E}
\left[
\min
\left(
\rho_t(\theta)A_t,\;
\operatorname{clip}
\left(
\rho_t(\theta),
1-\epsilon,
1+\epsilon
\right)A_t
\right)
\right],
\end{equation}
where $\rho_t(\theta)$ is the policy likelihood ratio, $A_t$ denotes the advantage estimate, and $\epsilon$ is the clipping parameter.

SAC-Online follows an off-policy procedure in which transitions generated by the current policy are continually added to its replay buffer and sampled during training. SAC learns a stochastic policy by maximizing expected return while encouraging entropy to promote exploration. Its objective augments the expected action value with an entropy regularization term:
\begin{equation}
\mathcal{L}_{\text{SAC}}
=
\mathbb{E}
\left[
Q(s,a)-\alpha\log\pi(a\mid s)
\right],
\end{equation}
where $\alpha$ controls the trade-off between reward maximization and policy entropy.

\subsubsection{Offline Algorithms}
\label{sec:offline_algo}

In the offline setting, TD3-BC, CQL-BC, and SAC-Offline perform policy optimization using identical fixed transition datasets generated through interaction with the personalized simulation environment. The transition dataset is constructed before policy optimization and remains unchanged throughout training. No additional simulator interaction occurs while the actor and critic networks are updated.

TD3-BC augments the deterministic policy-gradient objective with a behavior-cloning regularizer that constrains the learned policy toward the actions represented in the replay buffer:
\begin{equation}
\mathcal{L}_{\text{actor}}^{\text{TD3-BC}}
=
-\lambda\,
\mathbb{E}_{s\sim\mathcal{D}}
\left[
Q(s,\pi(s))
\right]
+
\mathbb{E}_{(s,a)\sim\mathcal{D}}
\left[
\|\pi(s)-a\|^2
\right],
\end{equation}
where $\mathcal{D}$ denotes the fixed replay buffer and $\lambda$ balances return maximization and adherence to the actions contained in the dataset. In GUIDE, these actions are produced by the randomized behavior policy used during buffer generation rather than by a historical patient policy.

CQL-BC follows the same actor--critic formulation as TD3-BC while augmenting the critic objective with a conservative regularization term. This discourages high Q-values for actions outside the support of the fixed replay buffer and reduces overestimation of out-of-distribution actions. CQL-BC also retains the behavior-cloning constraint, which regularizes its policy toward the actions generated by the randomized buffer-collection policy.

SAC-Offline applies the entropy-regularized SAC objective to the same fixed transition dataset used by TD3-BC and CQL-BC. Unlike SAC-Online, its replay buffer is not updated during policy optimization, and the algorithm performs no additional simulator interaction after the buffer has been constructed.

\section{Experimental Setup}
\label{sec:experimental_setup}

The AZT1D dataset\footnote{\url{https://data.mendeley.com/datasets/gk9m674wcx/1}}~\cite{khamesian2025azt1d}, the GUIDE source code\footnote{\url{https://github.com/SamanKhamesian/GUIDE}}, and the supplementary material\footnote{\href{https://github.com/SamanKhamesian/GUIDE/blob/main/material/supplementary_material.pdf}{Supplementary material on GitHub}}~\cite{guide_supplement} used to reproduce the experiments are publicly available.

\subsection{Dataset}
\label{sec:dataset}

This study uses the AZT1D dataset~\cite{khamesian2025azt1d} as the primary dataset for framework development and evaluation, and the OhioT1DM dataset~\cite{marling2020ohiot1dm} for external validation.

For each dataset, the data were sorted chronologically for each subject to preserve the natural temporal structure of glucose and behavioral dynamics, without random shuffling. The first 80\% of the time-ordered data was used to train the personalized glucose prediction model, while the remaining 20\% was reserved for the RL experiments. The RL portion was further divided chronologically, with the first 80\% used for agent training and the remaining 20\% used for evaluation. This separation prevents temporal information leakage between glucose-predictor training, RL training, and evaluation. Initial states were constructed using 6-hour sliding windows shifted forward by 1 hour, resulting in approximately 83.3\% overlap between adjacent windows.

\subsection{Training Configuration}
\label{sec:training_config}

We considered both offline and online RL settings. For each subject, offline data generation and online training used the same 100 initial states derived from the RL training subset and the same 24-step episode length. All algorithms used mini-batches of 256 transitions and five independent random seeds. The same bolus-safeguard configuration was applied across all algorithms during both training and evaluation, ensuring that all environment transitions were produced from executed safeguarded actions.

For each subject and random seed, the offline replay buffer was generated by initializing the simulator from the 100 RL training states. Each state was used for 20 independent 24-step rollouts, producing $48{,}000$ transitions per subject and seed. A randomized behavior policy selected the action type by applying \(\operatorname{argmax}\) to three independent scores sampled from a standard Gaussian distribution. Because the scores were identically distributed, each action type had equal selection probability. Carbohydrate and bolus insulin amounts were sampled uniformly over the experimental bounds \([c_{\min},c_{\max}]=[5,50]\)~g and \([i_{\min},i_{\max}]=[2,15]\)~U, respectively, while action times were sampled uniformly from the 12 five-minute positions within the upcoming hour. The symmetric distribution gave each action type equal selection probability, while the uniform sampling covered the complete predefined ranges of the continuous action components.

TD3-BC, CQL-BC, and SAC-Offline were trained on identical fixed transition datasets for each subject and seed. Each offline algorithm performed $10{,}000$ gradient-update steps by uniformly sampling the fixed replay buffer, with no additional simulator interaction or buffer modification. In contrast, PPO and SAC-Online were trained through direct simulator interaction. PPO collected on-policy trajectories for subsequent policy updates, whereas SAC-Online continuously added policy-generated transitions to its replay buffer and sampled them during training. Both online algorithms were trained for 20 epochs per subject.

The hyperparameter configurations for RL algorithms were consistent with those commonly reported in prior RL studies and were sufficient for stable training without extensive tuning~\cite{hettiarachchi2024g2p2c,emerson2023offline,marchetti2025deep}. Because neither the AZT1D nor the OhioT1DM dataset provides patient-specific insulin-action duration, insulin sensitivity factor, or insulin-to-carbohydrate ratio, the bolus-safeguard hyperparameters were determined using historical bolus information together with assumptions from prior simulation-based studies~\cite{emerson2023offline,lim2021blood}.

\vspace{-5pt}
\subsection{Evaluation Metrics}
\label{sec:evaluation_metrics}

\subsubsection{Glycemic Evaluation Metrics}
\label{sec:gly_metrics}

To evaluate the clinical relevance of the proposed RL framework, we assess glycemic control using standardized CGM-derived metrics recommended in clinical guidelines (Table~\ref{tab:glycemic_metrics}). These metrics jointly capture the safety and stability of glucose regulation. Time-in-Range (TIR) reflects overall glycemic control, Time-Below-Range (TBR) measures hypoglycemia exposure, Time-Above-Range (TAR) quantifies hyperglycemia burden, and the Coefficient of Variation (CV) assesses glucose variability.

\begin{table}[!t]
\centering
\footnotesize
\renewcommand{\arraystretch}{1.2}
\caption{Glycemic control metrics used for evaluation with clinically recommended targets \cite{battelino2019clinical, danne2017international}.}
\label{tab:glycemic_metrics}
\begin{tabularx}{\columnwidth}{l X l}
\toprule
Metric & Description & Target \\
\midrule
TIR 
& Percentage of time glucose is within 70--180 mg/dL 
& $>70\%$ \\
TBR
& Percentage of time glucose is below 70 mg/dL 
& $<4\%$ \\
TAR
& Percentage of time glucose exceeds 180 mg/dL 
& $<25\%$ \\
CV 
& Glucose variability
& $<36\%$ \\
\bottomrule
\end{tabularx}
\end{table}

\subsubsection{Behavioral Similarity Metrics}
\label{sec:behavrioal_metrics}

In addition to glycemic performance, we evaluate whether the behavioral patterns generated by the GUIDE framework resemble historical patient behavior. While TIR, TBR, and TAR quantify glycemic outcomes, behavioral alignment provides complementary insight into whether the learned policy produces realistic and clinically plausible action patterns. This evaluation is particularly valuable in decision-support systems, where large deviations from habitual behavior may reduce interpretability and practical adoption.

Let $x_s \in \mathbb{R}^n$ denote the behavioral profile generated by GUIDE for subject $s$, and let $y_s \in \mathbb{R}^n$ denote the corresponding historical patient profile. The behavioral profile and its feature components are described in Section~\ref{sec:behavioral_action_profile_analysis}. Because the behavioral features have different units and numerical scales, each feature is normalized using its mean historical value across the included subjects. For feature $i$, the cohort reference value is defined as
\begin{equation} \mu_i = \frac{1}{S}\sum_{s=1}^{S} y_{s,i}, \end{equation}
where $S$ denotes the number of included subjects. The normalized agent and patient features are then defined as
\begin{equation} \tilde{x}_{s,i} = \frac{x_{s,i}}{|\mu_i|}, \qquad \tilde{y}_{s,i} = \frac{y_{s,i}}{|\mu_i|}. \end{equation}
Thus, the same cohort-derived scaling factor is applied to the agent-generated and historical patient values of each feature. For clarity, the subject index $s$ is omitted from the following metric definitions.

\textit{\textbf{Cosine Similarity:}}
Cosine similarity evaluates the angular alignment between the normalized behavioral profiles. Values closer to 1 indicate stronger similarity in the relative distribution of behavioral features:
\begin{equation} \operatorname{CosSim}(\tilde{x},\tilde{y}) = \frac{\tilde{x}\cdot\tilde{y}}{\|\tilde{x}\|_2\|\tilde{y}\|_2}. \end{equation}

\textit{\textbf{Mean Relative Difference (MRD):}}
MRD measures the average relative deviation between corresponding behavioral features. Lower values indicate closer relative agreement between the agent-generated and historical patient profiles:
\begin{equation} \operatorname{MRD}(\tilde{x},\tilde{y}) = \frac{1}{n}\sum_{i=1}^{n}\left|\frac{\tilde{x}_i-\tilde{y}_i}{\tilde{y}_i}\right|. \end{equation}

\textit{\textbf{Cohort-Normalized $L_1$ Distance:}}
The cohort-normalized $L_1$ distance measures the average absolute difference between corresponding normalized behavioral features. Lower values indicate closer agreement in feature magnitude relative to the cohort reference scales:
\begin{equation} d_{L_1}(\tilde{x},\tilde{y}) = \frac{1}{n}\left\|\tilde{x}-\tilde{y}\right\|_1. \end{equation}

In the implementation, a small constant $\epsilon=10^{-6}$ is added to each potentially zero denominator in the cohort normalization, cosine similarity, and MRD calculations to prevent division by zero and ensure numerical stability. Together, these metrics characterize complementary aspects of behavioral alignment: cosine similarity captures directional structure, MRD measures subject-relative feature deviation, and the cohort-normalized $L_1$ distance quantifies the average scale-adjusted difference in feature magnitude.

\section{Results}
\label{sec:results}

\subsection{GLIMMER Performance Evaluation}
\label{sec:glimmer_performance}

\subsubsection{Prediction Performance}
\label{sec:glimmer_prediction_performance}

We first report the performance of GLIMMER, the glucose dynamics model used within the GUIDE simulation environment, which was independently evaluated on the AZT1D and OhioT1DM datasets using a 60-minute prediction horizon~\cite{khamesian2026glycemic}. The results show consistent predictive performance across the two datasets, with RMSE values of 22.48 and 23.97~mg/dL, MAE values of 15.58 and 15.83~mg/dL, and dysglycemia F1-scores of 82.07\% and 91.60\%, respectively.

\begin{table}[htbp]
\centering
\caption{Prediction performance of GLIMMER on the AZT1D and OhioT1DM datasets using a 60-minute prediction horizon. The complete evaluation results are reported in~\cite{khamesian2026glycemic}.}
\label{tab:glimmer_performance}
\begin{tabularx}{\columnwidth}{Xccc}
\toprule
\textbf{Dataset}
& \cellcolor{guidegreencell}\textbf{RMSE (mg/dL)}
& \cellcolor{guidepeachcell}\textbf{MAE (mg/dL)}
& \cellcolor{guidebluecell}\textbf{Dysglycemia F1 (\%)} \\
\midrule
AZT1D
& $22.48 \pm 3.57$
& $15.58 \pm 2.87$
& 82.07 \\
OhioT1DM
& $23.97 \pm 3.77$
& $15.83 \pm 2.09$
& 91.60 \\
\bottomrule
\end{tabularx}
\end{table}

\subsubsection{Counterfactual Response Evaluation}
\label{sec:glimmer_counterfactual}

We next examined whether GLIMMER produced consistent short-term responses when insulin and carbohydrate inputs were altered from matched held-out states. GLIMMER was trained independently for each of the 25 AZT1D patients using five random seeds. For each patient and seed, insulin counterfactuals of 0, 2, 5, and 10~U and carbohydrate counterfactuals of 0, 10, 30, and 50~g were applied to the same 10 patient-specific test states, resulting in 250 patient--state evaluations per seed. Each nonzero intervention was compared with its matched zero-action prediction. Mean $\Delta G$ denotes the average difference over the 5--60-minute predicted trajectory, while $\Delta G_{60}$ denotes the difference at the 60-minute horizon.

\begin{figure}[t]
\centering
\includegraphics[width=\columnwidth]{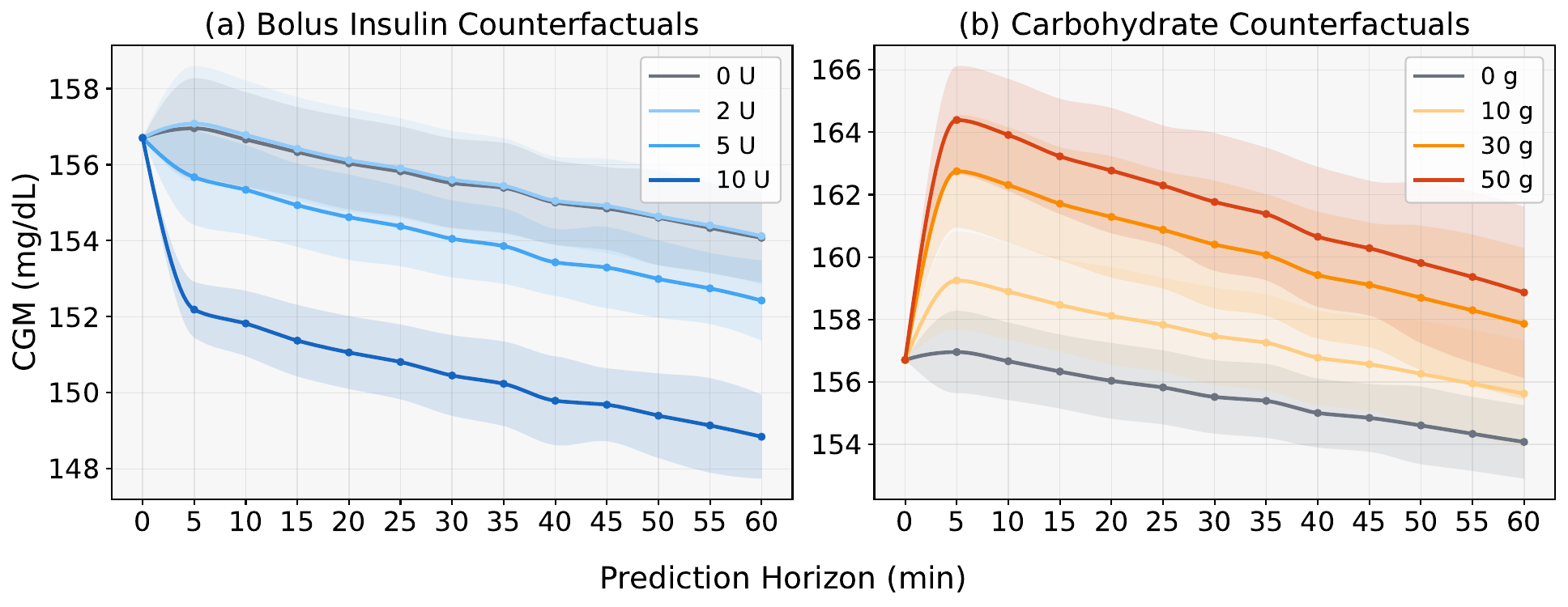}
\caption{GLIMMER counterfactual glucose trajectories under increasing bolus-insulin and carbohydrate inputs. For each seed, predictions were averaged across the 10 test states within each patient and subsequently across the 25 patients. Solid lines represent the mean of the five seeds, and shaded regions indicate $\pm1$ SD across seeds.}
\label{fig:glimmer_counterfactual}
\end{figure}

\begin{table}[t]
\centering
\caption{Counterfactual response magnitude relative to the matched zero-action prediction. Values are cohort mean $\pm$ SD across patient-level averages after averaging the five seeds.}
\label{tab:glimmer_counterfactual_magnitude}
\begin{tabularx}{\columnwidth}{lXcc}
\toprule
\textbf{Intervention}
& \textbf{Level}
& \cellcolor{guidepeachcell}\textbf{Mean $\Delta G$ (mg/dL)}
& \cellcolor{guidebluecell}\textbf{$\Delta G_{60}$ (mg/dL)} \\
\midrule
Insulin      & 2 U  & $-0.24 \pm 4.52$  & $-0.35 \pm 4.07$  \\
Insulin      & 5 U  & $-1.49 \pm 9.96$  & $-1.65 \pm 8.97$  \\
Insulin      & 10 U & $-5.07 \pm 15.00$ & $-5.23 \pm 13.74$ \\
\midrule
Carbohydrate & 10 g & $1.91 \pm 2.08$   & $1.55 \pm 2.14$   \\
Carbohydrate & 30 g & $4.77 \pm 6.36$   & $3.78 \pm 6.38$   \\
Carbohydrate & 50 g & $6.09 \pm 9.73$   & $4.79 \pm 9.71$   \\
\bottomrule
\end{tabularx}
\end{table}

\begin{table}[htbp]
\centering
\caption{Directional and complete-ranking consistency of GLIMMER counterfactual responses (mean $\pm$ SD across five seeds). Directional consistency is calculated from 750 nonzero-versus-zero comparisons per seed, and complete-ranking consistency from 250 patient--state evaluations per seed.}
\label{tab:glimmer_counterfactual_consistency}
\begin{tabularx}{\columnwidth}{Xcc}
\toprule
\textbf{Expected ordering}
& \cellcolor{guidepeachcell}\textbf{Directional}
& \cellcolor{guidebluecell}\textbf{Complete ranking} \\
\midrule
$0~\mathrm{U} > 2~\mathrm{U} > 5~\mathrm{U} > 10~\mathrm{U}$
& $82.9\% \pm 3.7\%$
& $71.3\% \pm 4.8\%$ \\
$0~\mathrm{g} < 10~\mathrm{g} < 30~\mathrm{g} < 50~\mathrm{g}$
& $85.4\% \pm 4.6\%$
& $77.7\% \pm 3.3\%$ \\
\bottomrule
\end{tabularx}
\end{table}

As shown in Fig.~\ref{fig:glimmer_counterfactual} and Table~\ref{tab:glimmer_counterfactual_magnitude}, increasing bolus insulin produced progressively larger reductions in predicted glucose at the cohort level, whereas increasing carbohydrate intake produced progressively larger increases. The mean 5--60-minute response changed from $-0.24$~mg/dL at 2~U to $-5.07$~mg/dL at 10~U, while the corresponding carbohydrate response increased from 1.91~mg/dL at 10~g to 6.09~mg/dL at 50~g. Across the evaluated intervention levels, the cross-seed SD ranged from 0.47 to 1.65~mg/dL for mean $\Delta G$ and from 0.57 to 1.71~mg/dL for $\Delta G_{60}$. These values were substantially smaller than the cross-patient SDs, indicating reproducible cohort-level responses across independent GLIMMER training runs while retaining heterogeneous patient-specific responses.

Using the mean predicted CGM over the 5--60-minute trajectory, Table~\ref{tab:glimmer_counterfactual_consistency} summarizes the directional and complete-ranking consistency of the counterfactual responses. Directional consistency exceeded 82\% for both interventions, while complete dose-level ordering exceeded 71\% across the evaluated patient-specific states. These results indicate that GLIMMER generally preserves the expected direction and dose ordering of insulin and carbohydrate responses within the states evaluated in GUIDE.

\subsubsection{Recursive Rollout Performance}
\label{sec:glimmer_rollout}

To evaluate error accumulation during recursive simulation, GLIMMER was initialized for each of the 25 AZT1D patients using six hours of held-out observations and rolled forward for 24 consecutive hourly steps. Predicted CGM values were recursively incorporated into subsequent inputs, while the recorded carbohydrate intake, bolus insulin, and basal insulin inputs were replayed. Table~\ref{tab:glimmer_rollout} reports the hourly and cumulative prediction errors at selected rollout horizons.

\begin{table}[t]
\centering
\caption{GLIMMER prediction error during the recursive 24-hour rollout at selected horizons $H$, where $H$ denotes the rollout hour. Hourly metrics are calculated from the 12 predictions within hour $H$, whereas cumulative metrics include all predictions from hour 1 through $H$. Values are reported as mean $\pm$ standard deviation across 25 patients.}
\label{tab:glimmer_rollout}
\resizebox{\columnwidth}{!}{
\begin{tabular}{ccccc}
\toprule
\multirow{2}{*}{\textbf{$H$ (h)}} &
\multicolumn{2}{c}{\cellcolor{guidegreencell}\textbf{Hourly Error (mg/dL)}} &
\multicolumn{2}{c}{\cellcolor{guidepurplecell}\textbf{Cumulative Error (mg/dL)}} \\
\cmidrule(lr){2-3}\cmidrule(lr){4-5}
& \cellcolor{guidepeachcell}\textbf{MAE}
& \cellcolor{guidebluecell}\textbf{RMSE}
& \cellcolor{guidepeachcell}\textbf{MAE}
& \cellcolor{guidebluecell}\textbf{RMSE} \\
\midrule
1  & $14.16 \pm 3.25$ & $21.20 \pm 4.76$  & $14.16 \pm 3.25$ & $21.20 \pm 4.76$ \\
6  & $20.56 \pm 5.47$ & $29.22 \pm 6.85$  & $18.48 \pm 4.50$ & $27.37 \pm 5.84$ \\
12 & $21.45 \pm 6.12$ & $30.76 \pm 7.41$  & $19.41 \pm 5.05$ & $28.49 \pm 6.58$ \\
18 & $22.50 \pm 6.71$ & $31.50 \pm 8.77$  & $20.77 \pm 6.10$ & $29.92 \pm 7.44$ \\
24 & $22.79 \pm 8.36$ & $32.90 \pm 10.46$ & $20.91 \pm 8.56$ & $31.12 \pm 9.68$ \\
\bottomrule
\end{tabular}}
\end{table}

As shown in Table~\ref{tab:glimmer_rollout}, the largest increase in prediction error occurred during the first six hours, after which both hourly and cumulative errors increased more gradually. By hour 6, the six-hour CGM history supplied to GLIMMER consisted entirely of recursively predicted values. From hour 6 to hour 24, cumulative MAE and RMSE increased by only $2.43$ and $3.75$~mg/dL, respectively, remaining within a comparable range. These results indicate that GLIMMER maintains reasonable predictive stability throughout the 24-hour GUIDE rollout.

\subsection{GUIDE Framework Evaluation}
\label{sec:guide_evaluation}

\subsubsection{Comparative Glycemic Performance}
\label{sec:glycemic_performance}

\begin{figure*}[t]
\small
\centering
\includegraphics[width=0.9\linewidth]{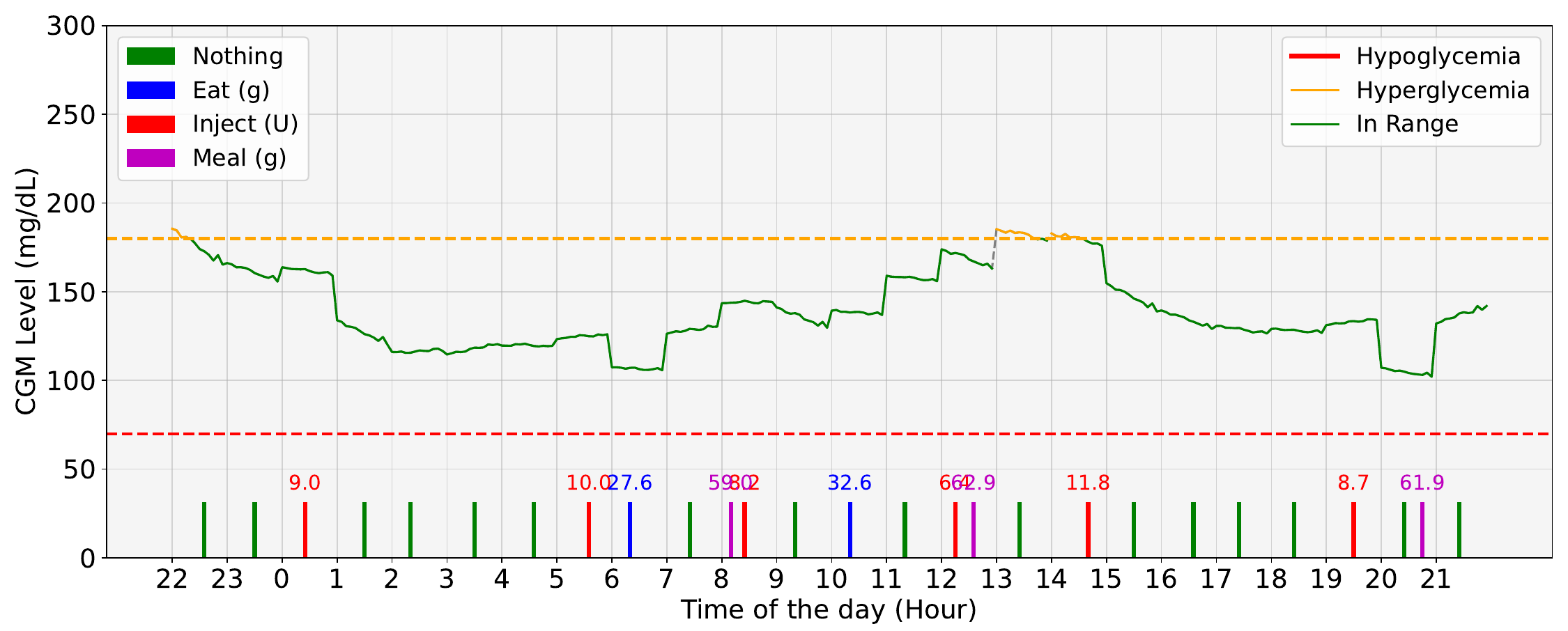}
\caption{Representative full-day simulation under full adherence using the personalized glucose prediction model. The solid green curve shows the predicted glucose trajectory across 24 decision steps. The dashed lines at 70 and 180 mg/dL mark the hypoglycemia and hyperglycemia thresholds, respectively, and bound the target range. Vertical markers denote decision and meal events: green indicates no intervention (Nothing), blue indicates carbohydrate intake recommended by the RL agent (Eat), red indicates a recommended bolus insulin dose (Inject), and magenta indicates a structured meal generated by the human-inspired meal controller. Numerical labels report carbohydrate amounts in grams for Eat and Meal events and insulin doses in units for Inject events.}
\label{fig:simulation}
\end{figure*}

Glycemic performance was evaluated using personalized glucose prediction models trained separately on each individual's historical data from the AZT1D and OhioT1DM datasets. For each subject in both datasets, every algorithm was assessed across 10 randomized initial states and 5 different random seeds, with each roll-out spanning 24 decision steps corresponding to a full simulated day. The reported metrics summarize glycemic outcomes averaged over these repeated daily simulations for each patient. Fig.~\ref{fig:simulation} shows a representative full-day simulation, including the glucose level trajectory and corresponding action events. All evaluations assume full adherence, where recommended actions are executed exactly within the simulation environment, enabling direct comparison of policy outcomes.

\begin{table*}[t]
\centering
\small
\caption{Performance comparison of RL algorithms within the proposed \textbf{GUIDE} framework under identical environment and reward settings across the AZT1D and OhioT1DM datasets. Results are reported for offline algorithms (TD3-BC, CQL-BC, SAC-Offline) and online algorithms (PPO, SAC-Online). Metrics TIR, TAR, TBR, and CV are reported as mean $\pm$ standard deviation across 25 subjects for AZT1D and 12 subjects for OhioT1DM. The \textbf{Random} policy illustrates outcomes under unstructured action selection within the simulation environment, while \textbf{History} summarizes each subject's historical glycemic outcomes for contextual reference.}
\begin{tabularx}{\textwidth}{Xcccccc}
\toprule
\textbf{Algorithm} & \cellcolor{guidegreencell}\textbf{TIR (\%) $\uparrow$} & \cellcolor{guidepeachcell}\textbf{TAR (\%) $\downarrow$} & \cellcolor{guidebluecell}\textbf{TBR (\%) $\downarrow$} & \cellcolor{guidepurplecell}\textbf{CV (\%) $\downarrow$} & \cellcolor{guideyellowcell}\textbf{Policy Type} & \cellcolor{guidemintcell}\textbf{Interaction Mode} \\
\midrule

\multicolumn{7}{c}{\textit{AZT1D Dataset}} \\
\midrule
TD3-BC      & \(81.64 \pm 20.59\)           & \(17.83 \pm 20.69\)           & \(0.53 \pm 1.27\)           & \(15.49 \pm 3.74\)           & off-policy & offline \\
CQL-BC      & \(\mathbf{84.18 \pm 19.89}\)  & \(\mathbf{15.39 \pm 19.99}\)  & \(\mathbf{0.43 \pm 1.27}\)  & \(\mathbf{15.13 \pm 3.88}\)  & off-policy & offline \\
SAC-Offline & \(71.32 \pm 25.17\)           & \(27.59 \pm 25.95\)           & \(1.09 \pm 1.98\)           & \(16.86 \pm 4.19\)           & off-policy & offline \\
\midrule
PPO         & \(78.25 \pm 20.43\)           & \(21.06 \pm 20.70\)           & \(0.68 \pm 1.43\)           & \(15.32 \pm 3.79\)           & on-policy  & online  \\
SAC-Online  & \(70.55 \pm 25.09\)           & \(28.35 \pm 25.72\)           & \(1.10 \pm 2.02\)           & \(16.85 \pm 4.11\)           & off-policy & online  \\
\midrule
Random      & \(61.75 \pm 26.36\)           & \(31.54 \pm 30.08\)           & \(5.88 \pm 11.68\)          & \(20.74 \pm 8.40\)           & \(-\)      & \(-\)   \\
History     & \(77.41 \pm 11.47\)           & \(21.10 \pm 12.01\)           & \(1.48 \pm 1.53\)           & \(29.70 \pm 4.10\)           & \(-\)      & \(-\)   \\

\midrule
\multicolumn{7}{c}{\textit{OhioT1DM Dataset}} \\
\midrule
TD3-BC      & \(75.73 \pm 11.30\) & \(21.04 \pm 9.30\) & \(3.23 \pm 4.31\) & \(\mathbf{24.74 \pm 7.36}\) & off-policy & offline \\
CQL-BC      & \(\mathbf{77.64 \pm 8.81}\)  & \(\mathbf{19.55 \pm 8.53}\) & \(\mathbf{2.81 \pm 3.56}\) & \(25.04 \pm 8.79\) & off-policy & offline \\
SAC-Offline & \(61.90 \pm 12.48\) & \(35.04 \pm 13.66\) & \(3.06 \pm 4.39\) & \(29.19 \pm 7.38\) & off-policy & offline \\
\midrule
PPO         & \(68.11 \pm 8.71\)  & \(27.92 \pm 7.44\) & \(3.97 \pm 4.78\) & \(24.44 \pm 8.71\) & on-policy & online \\
SAC-Online  & \(61.70 \pm 13.49\) & \(34.59 \pm 13.79\) & \(2.88 \pm 3.95\) & \(28.68 \pm 6.92\) & off-policy & online \\
\midrule
Random & \(54.34 \pm 17.14\) & \(39.94 \pm 19.04\) & \(4.89 \pm 5.78\) & \(29.91 \pm 7.71\) & \(-\) & \(-\) \\
History     & \(63.62 \pm 10.31\) & \(32.98 \pm 11.47\) & \(3.40 \pm 2.39\) & \(36.66 \pm 3.80\) & \(-\) & \(-\) \\
\bottomrule
\end{tabularx}
\label{tab:glycemic_comparison}
\end{table*}

\begin{table*}[t]
\centering
\footnotesize
\caption{Pairwise comparison of algorithms based on TIR$^{1}$ across the AZT1D and OhioT1DM datasets. Values represent Holm--Bonferroni adjusted p-values obtained from Wilcoxon signed-rank tests using per-subject TIR across 25 subjects for AZT1D and 12 subjects for OhioT1DM. The correction controls the family-wise error rate across the 15 pairwise comparisons within each dataset. Bold values denote statistically significant differences ($p < 0.05$) after adjustment.}
\begin{tabularx}{\textwidth}{l*{6}{>{\centering\arraybackslash}X}}
\toprule
Algorithm
& \cellcolor{guidegreencell}\textbf{TD3-BC}
& \cellcolor{guidepeachcell}\textbf{CQL-BC}
& \cellcolor{guidebluecell}\textbf{PPO}
& \cellcolor{guidepurplecell}\textbf{SAC-Offline}
& \cellcolor{guideyellowcell}\textbf{SAC-Online}
& \cellcolor{guidemintcell}\textbf{Random} \\
\toprule

\multicolumn{7}{c}{\textit{AZT1D Dataset}} \\
\midrule

\cellcolor{guidegreencell}\textbf{TD3-BC}
& --
& $\mathbf{4.20\times10^{-2}}$
& $\mathbf{6.48\times10^{-3}}$
& $\mathbf{8.94\times10^{-7}}$
& $\mathbf{2.15\times10^{-6}}$
& $\mathbf{1.23\times10^{-4}}$ \\

\cellcolor{guidepeachcell}\textbf{CQL-BC}
& $\mathbf{4.20\times10^{-2}}$
& --
& $\mathbf{1.23\times10^{-4}}$
& $\mathbf{1.55\times10^{-6}}$
& $\mathbf{8.94\times10^{-7}}$
& $\mathbf{1.23\times10^{-4}}$ \\

\cellcolor{guidebluecell}\textbf{PPO}
& $\mathbf{6.48\times10^{-3}}$
& $\mathbf{1.23\times10^{-4}}$
& --
& $\mathbf{1.87\times10^{-4}}$
& $\mathbf{2.82\times10^{-5}}$
& $\mathbf{1.50\times10^{-3}}$ \\

\cellcolor{guidepurplecell}\textbf{SAC-Offline}
& $\mathbf{8.94\times10^{-7}}$
& $\mathbf{1.55\times10^{-6}}$
& $\mathbf{1.87\times10^{-4}}$
& --
& $7.05\times10^{-1}$
& $6.91\times10^{-1}$ \\

\cellcolor{guideyellowcell}\textbf{SAC-Online}
& $\mathbf{2.15\times10^{-6}}$
& $\mathbf{8.94\times10^{-7}}$
& $\mathbf{2.82\times10^{-5}}$
& $7.05\times10^{-1}$
& --
& $7.05\times10^{-1}$ \\

\cellcolor{guidemintcell}\textbf{Random}
& $\mathbf{1.23\times10^{-4}}$
& $\mathbf{1.23\times10^{-4}}$
& $\mathbf{1.50\times10^{-3}}$
& $6.91\times10^{-1}$
& $7.05\times10^{-1}$
& -- \\

\midrule
\multicolumn{7}{c}{\textit{OhioT1DM Dataset}} \\
\midrule

\cellcolor{guidegreencell}\textbf{TD3-BC}
& --
& $7.61\times10^{-1}$
& $\mathbf{1.61\times10^{-2}}$
& $\mathbf{2.20\times10^{-2}}$
& $\mathbf{2.20\times10^{-2}}$
& $\mathbf{7.32\times10^{-3}}$ \\

\cellcolor{guidepeachcell}\textbf{CQL-BC}
& $7.61\times10^{-1}$
& --
& $\mathbf{1.61\times10^{-2}}$
& $\mathbf{1.27\times10^{-2}}$
& $\mathbf{1.27\times10^{-2}}$
& $\mathbf{7.32\times10^{-3}}$ \\

\cellcolor{guidebluecell}\textbf{PPO}
& $\mathbf{1.61\times10^{-2}}$
& $\mathbf{1.61\times10^{-2}}$
& --
& $2.56\times10^{-1}$
& $2.56\times10^{-1}$
& $\mathbf{4.64\times10^{-2}}$ \\

\cellcolor{guidepurplecell}\textbf{SAC-Offline}
& $\mathbf{2.20\times10^{-2}}$
& $\mathbf{1.27\times10^{-2}}$
& $2.56\times10^{-1}$
& --
& $9.10\times10^{-1}$
& $\mathbf{2.39\times10^{-2}}$ \\

\cellcolor{guideyellowcell}\textbf{SAC-Online}
& $\mathbf{2.20\times10^{-2}}$
& $\mathbf{1.27\times10^{-2}}$
& $2.56\times10^{-1}$
& $9.10\times10^{-1}$
& --
& $\mathbf{2.39\times10^{-2}}$ \\

\cellcolor{guidemintcell}\textbf{Random}
& $\mathbf{7.32\times10^{-3}}$
& $\mathbf{7.32\times10^{-3}}$
& $\mathbf{4.64\times10^{-2}}$
& $\mathbf{2.39\times10^{-2}}$
& $\mathbf{2.39\times10^{-2}}$
& -- \\

\bottomrule
\end{tabularx}
\par\vspace{5pt}
\noindent\begin{minipage}{\textwidth}
\footnotesize
\textsuperscript{1}\,History is retained as a descriptive reference in Table~\ref{tab:glycemic_comparison} but excluded from the inferential comparisons reported here.
\end{minipage}
\label{tab:p_value}
\vspace{-10pt}
\end{table*}

We calculated aggregate glycemic metrics for the 25 AZT1D subjects and 12 OhioT1DM subjects, including TIR, TAR, TBR, and CV (Table~\ref{tab:glycemic_comparison}). On the AZT1D dataset, CQL-BC achieved the highest mean TIR (\(84.18 \pm 19.89\%\)), followed by TD3-BC (\(81.64 \pm 20.59\%\)). These algorithms also exhibited the lowest mean TAR and TBR among the evaluated RL controllers. PPO ranked next in TIR, while SAC-Offline slightly outperformed SAC-Online. The Random policy, included to represent unstructured action selection within the same simulation environment, yielded lower TIR and higher TBR than the learned policies. The History row reports each subject's historical glycemic outcomes under real-world conditions and is provided for contextual comparison rather than direct equivalence, since historical data and simulated evaluations are generated under different settings. CQL-BC also achieved the best performance on the OhioT1DM dataset, with a mean TIR of \(77.64 \pm 8.81\%\), TAR of \(19.55 \pm 8.53\%\), TBR of \(2.81 \pm 3.56\%\), and CV of \(25.04 \pm 8.79\%\).

To assess whether the observed differences in TIR among the algorithms were statistically significant, pairwise Wilcoxon signed-rank tests were performed using per-subject TIR values for each dataset (Table~\ref{tab:p_value}). This nonparametric test compares paired TIR values without assuming that their differences are normally distributed. To account for multiple comparisons, the Holm--Bonferroni procedure was applied separately within each dataset across all 15 pairwise comparisons to control the family-wise error rate~\cite{preisser2011multiple, lee2018proper}.

\subsubsection{Behavioral Action Profile Analysis}
\label{sec:behavioral_action_profile_analysis}

The objective of this analysis is to evaluate the behavioral plausibility of the actions generated by the RL agent. While glycemic metrics assess clinical effectiveness, they do not indicate whether the agent’s action patterns remain realistic and consistent with human behavior. Therefore, beyond outcome-based evaluation, we examine whether the learned policy produces action frequencies, magnitudes, and temporal patterns comparable to the patient’s historical behavior. This analysis allows us to assess whether the generated actions remain clinically plausible and behaviorally interpretable.

To enable a direct comparison, we summarize action trajectories using a compact behavioral feature vector for each subject. This vector comprises seven features: daily bolus injection frequency, total daily bolus insulin, daily meal frequency, total daily carbohydrate intake, bolus-to-carbohydrate ratio, the average time gap between consecutive eating actions (meal gap), and the average time gap between consecutive bolus injections (bolus gap). Together, these features capture action frequency, magnitude, and temporal spacing, providing an interpretable behavioral signature for each subject.

\begin{table}[t]
\centering
\small
\renewcommand{\arraystretch}{1.1}
\caption{Aggregate behavioral similarity metrics for the CQL-BC algorithm across the AZT1D and OhioT1DM datasets. The AZT1D results include 24 subjects, with subject \#23 excluded because of missing values, while the OhioT1DM results include all 12 subjects.}
\begin{tabularx}{\columnwidth}{Xccc}
\toprule
\textbf{Statistic}
& \cellcolor{guidegreencell}\textbf{Cosine Similarity $\uparrow$}
& \cellcolor{guidepeachcell}\textbf{MRD $\downarrow$}
& \cellcolor{guidebluecell}\textbf{$d_{L_1}$ $\downarrow$} \\
\midrule

\multicolumn{4}{c}{\textit{AZT1D Dataset}} \\
\midrule
Mean   & 0.767 & 0.794 & 0.667 \\
STD    & 0.128 & 0.261 & 0.208 \\
Median & 0.779 & 0.799 & 0.663 \\
\midrule

\multicolumn{4}{c}{\textit{OhioT1DM Dataset}} \\
\midrule
Mean   & 0.774 & 0.840 & 0.712 \\
STD    & 0.162 & 0.381 & 0.313 \\
Median & 0.805 & 0.816 & 0.651 \\
\bottomrule
\end{tabularx}
\label{tab:cql_bc_metrics}
\end{table}

For each subject, the patient behavioral vector is computed from one month of historical training data, while the agent vector is averaged across 10 evaluation runs for each of 5 random seeds. To quantify behavioral alignment across the cohorts, we use cosine similarity, MRD, and cohort-normalized $L_1$ distance, as defined in Section~\ref{sec:evaluation_metrics}, with the aggregate results reported in Table~\ref{tab:cql_bc_metrics}. CQL-BC achieved mean cosine similarity, MRD, and cohort-normalized $L_1$ distance values of 0.767, 0.794, and 0.667 on AZT1D, respectively, compared with 0.774, 0.840, and 0.712 on OhioT1DM. These comparable results indicate consistent behavioral alignment across both datasets. Together, these results support the behavioral plausibility of the learned policy while showing that it adapts action frequency and magnitude rather than exactly reproducing historical behavior.

\subsection{Ablation Studies}
\label{sec:ablation_studies}

Based on the comparative results in Table~\ref{tab:glycemic_comparison}, we selected CQL-BC, which achieved the strongest overall glycemic performance among the evaluated algorithms, for the ablation studies and partial-adherence evaluation. Using the AZT1D dataset, these analyses involved 25 patients and 10 evaluation tests per patient. The experiments examined individual components of the GUIDE action space and reward function, as well as partial adherence. In each experiment, only the designated component or adherence condition was modified, while all other framework components and experimental settings, including the simulation environment, human-inspired meal generator, basal insulin generator, and bolus safeguard, were kept identical across conditions.

\vspace{5pt}
\subsubsection{Action Space Ablation}
\label{sec:action_space_ablation}

Five action-space configurations were considered. \textit{Complete} uses the complete action space, allowing the agent to select the action type, magnitude, and timing. In \textit{Fixed timing}, the carbohydrate and bolus insulin actions and their magnitudes are retained, but the timing component is removed and each selected action is applied at the first five-minute position of the upcoming hour. \textit{Carbohydrate only} removes bolus insulin from the agent's action space, allowing only carbohydrate recommendations, whereas \textit{Insulin only} removes carbohydrate recommendations and allows only bolus insulin actions. The \textit{Random} policy provides a reference in which actions are randomly selected from the complete action space under the same environment settings.

\begin{table}[t]
\centering
\footnotesize
\caption{Glycemic performance under different action-space configurations. Results are reported as mean $\pm$ standard deviation across 25 patients, with 10 evaluation tests per patient.}
\begin{tabularx}{\columnwidth}{Xccc}
\toprule
\textbf{Configuration}
& \cellcolor{guidegreencell}\textbf{TIR (\%) $\uparrow$}
& \cellcolor{guidepeachcell}\textbf{TBR (\%) $\downarrow$}
& \cellcolor{guidebluecell}\textbf{TAR (\%) $\downarrow$} \\
\midrule
Complete          & \(84.40 \pm 19.70\) & \(0.50 \pm 4.97\) & \(15.10 \pm 19.85\) \\
Fixed timing      & \(74.09 \pm 21.47\) & \(3.85 \pm 6.20\) & \(22.07 \pm 21.50\) \\
Carbohydrate only & \(71.00 \pm 21.26\) & \(1.49 \pm 5.12\) & \(27.51 \pm 21.50\) \\
Insulin only      & \(72.30 \pm 20.01\) & \(5.69 \pm 5.56\) & \(22.01 \pm 20.17\) \\
Random            & \(62.07 \pm 25.02\) & \(5.79 \pm 7.42\) & \(32.13 \pm 26.24\) \\
\bottomrule
\end{tabularx}
\label{tab:action_space_ablation}
\end{table}

As shown in Table~\ref{tab:action_space_ablation}, the \textit{Complete} configuration achieved the highest TIR and the lowest TBR and TAR among the evaluated conditions. Removing agent-proposed bolus insulin in the \textit{Carbohydrate only} condition increased TAR to 27.51\%, as the agent could no longer recommend corrective bolus actions in response to elevated glucose. Conversely, removing agent-proposed carbohydrate intake in the \textit{Insulin only} condition increased TBR to 5.69\%, as the agent could no longer recommend snacks in response to low glucose. The lower performance under \textit{Fixed timing} further demonstrates the importance of allowing the policy to optimize action timing, while the \textit{Random} policy produced the lowest TIR and the highest TBR and TAR.

\subsubsection{Reward Function Ablation}
\label{sec:reward_function_ablation}

The reward function ablation examines three principal clinical components: glycemic regulation, meal recommendation, and insulin recommendation. \textit{Complete} uses the complete reward function. \textit{No glycemic} removes both the step-wise and episode-level glycemic rewards, while \textit{No meal} and \textit{No insulin} remove the reward terms associated with meal and insulin decisions, respectively. The sleep and stability components were retained across all configurations because they function as behavioral regularizers that discourage unnecessary nighttime and repeated interventions, rather than defining the primary treatment objectives.

\begin{table}[t]
\centering
\footnotesize
\caption{Glycemic performance under different reward function configurations. Results are reported as mean $\pm$ standard deviation across 25 patients, with 10 evaluation tests per patient.}
\begin{tabularx}{\columnwidth}{Xccc}
\toprule
\textbf{Configuration}
& \cellcolor{guidegreencell}\textbf{TIR (\%) $\uparrow$}
& \cellcolor{guidepeachcell}\textbf{TBR (\%) $\downarrow$}
& \cellcolor{guidebluecell}\textbf{TAR (\%) $\downarrow$} \\
\midrule
Complete    & \(84.40 \pm 19.70\) & \(0.50 \pm 4.97\) & \(15.10 \pm 19.85\) \\
No glycemic & \(64.77 \pm 27.79\) & \(5.65 \pm 5.73\) & \(29.58 \pm 28.48\) \\
No meal     & \(74.81 \pm 23.88\) & \(3.60 \pm 5.86\) & \(21.59 \pm 23.05\) \\
No insulin  & \(73.84 \pm 22.54\) & \(1.90 \pm 6.29\) & \(24.26 \pm 22.15\) \\
\bottomrule
\end{tabularx}
\label{tab:reward_function_ablation}
\end{table}

As shown in Table~\ref{tab:reward_function_ablation}, the \textit{Complete} configuration achieved the highest TIR and the lowest TBR and TAR. Removing the glycemic component produced the largest performance reduction, decreasing TIR from 84.40\% to 64.77\% while increasing both TBR and TAR. Removing the meal component reduced TIR to 74.81\% and increased TBR to 3.60\%, demonstrating the importance of meal-related reward signals for learning appropriate carbohydrate recommendations in response to low glucose. Removing the insulin component reduced TIR to 73.84\% and increased TAR to 24.26\%, demonstrating the contribution of insulin-related reward signals to learning corrective bolus recommendations during elevated glucose. Overall, these results show that all three principal clinical reward components contribute to GUIDE's glycemic performance, with the glycemic component providing the largest overall contribution.

\subsubsection{Partial Adherence Analysis}
\label{sec:adherence_analysis}

To examine the effect of partial adherence, the same patient-specific policy was evaluated under adherence targets of 100\%, 75\%, and 50\%. Adherence was applied only to carbohydrate and bolus insulin recommendations generated by the agent. At each decision step, an \textsc{Eat} or \textsc{Inject} recommendation was independently followed according to the specified adherence probability; otherwise, it was replaced with \textsc{Nothing}. The human-inspired meal generator, glucose-responsive basal insulin generator, bolus safeguard, and all other experimental settings remained unchanged across adherence conditions. The evaluation results were first averaged within each patient, and Table~\ref{tab:adherence_evaluation} reports the mean and standard deviation across patients.

\begin{table}[htbp]
\centering
\footnotesize
\caption{Glycemic performance under different adherence targets. Results are reported as mean $\pm$ standard deviation across 25 patients, with 10 evaluation tests per patient.}
\begin{tabularx}{\columnwidth}{Xccc}
\toprule
\textbf{Adherence target}
& \cellcolor{guidegreencell}\textbf{TIR (\%) $\uparrow$}
& \cellcolor{guidepeachcell}\textbf{TBR (\%) $\downarrow$}
& \cellcolor{guidebluecell}\textbf{TAR (\%) $\downarrow$} \\
\midrule
\(100\%\) & \(84.40 \pm 19.70\) & \(0.50 \pm 4.97\) & \(15.10 \pm 19.85\) \\
\(75\%\)  & \(78.19 \pm 22.68\) & \(2.63 \pm 5.35\) & \(19.18 \pm 21.94\) \\
\(50\%\)  & \(68.45 \pm 24.37\) & \(4.97 \pm 6.16\) & \(26.58 \pm 25.08\) \\
\bottomrule
\end{tabularx}
\label{tab:adherence_evaluation}
\end{table}

As adherence decreased, glycemic performance deteriorated progressively. TIR declined from 84.40\% under full adherence to 78.19\% at the 75\% adherence target and 68.45\% at the 50\% target. Correspondingly, TBR increased from 0.50\% to 2.63\% and 4.97\%, while TAR increased from 15.10\% to 19.18\% and 26.58\%. These results indicate that reduced adherence, involving missed bolus and carbohydrate recommendations, increases exposure to both hyperglycemia and hypoglycemia. Therefore, maintaining adherence as close as possible to the recommended actions is desirable for achieving better glycemic management.

\section{Discussion}
\label{sec:discussion}

Interpretability in GUIDE is examined through the structure and resulting behavior of the learned policy. The action-space ablation quantifies the contributions of action type, magnitude, and timing, while the reward-function ablation evaluates the principal objectives guiding policy learning. Complementarily, the Behavioral Action Profile Analysis compares seven clinically and behaviorally relevant characteristics of the generated recommendations with the distributions observed in patients’ historical data using cosine similarity, mean relative difference, and cohort-normalized L1 distance. Together, these analyses provide cohort- and individual-level evidence regarding the contribution and behavioral plausibility of GUIDE’s recommendations.

Nevertheless, several limitations of the proposed framework should be considered. First, the framework assumes that recommended actions are followed at the specified decision times. In real-world settings, strict adherence to such recommendations is rarely feasible, as individuals may delay, ignore, or only partially follow suggested actions due to practical or behavioral constraints. Accordingly, the learned policy should not be interpreted as a prescriptive treatment protocol. Instead, it provides insight into how glycemic outcomes may evolve under higher levels of adherence and serves as a decision-support reference for exploring “what-if” behavioral scenarios. Modeling partial adherence and user acceptance of the decision-support system remains an open challenge.

Second, although the policy is trained using trajectories generated through interaction with the simulation environment, the diversity of these trajectories is constrained by the data-collection strategy used to populate the replay buffer. The CQL-BC agent can learn only from state–action regions that were sufficiently explored during this rollout phase. If certain behavioral patterns, glucose excursions, or action combinations were rarely encountered during trajectory generation in the simulator, the learned policy may demonstrate limited robustness in those regions. Consequently, performance depends on the diversity of trajectories generated by the exploration policy, and broader scenario coverage may improve policy generalization and stability.

Third, the quality of historical patient logs influences the accuracy of behavioral profiling. In some cases, food logs were incomplete or inconsistently recorded. Because behavioral feature vectors are derived from these historical logs, missing or inaccurate entries may reduce the precision of the patient reference profile. A more comprehensive and consistently recorded dataset would likely improve the fidelity of behavioral alignment analysis, although obtaining such data in real-world settings remains challenging due to variable patient engagement and adherence.

Although the reward terms were selected based on clinically established glycemic targets and clinically motivated behavioral considerations, and the ablation results demonstrate the contribution of the main components, the current formulation uses a linear combination with predefined coefficients. Future work will investigate reward functions aligned with patient and clinician preferences in real-world studies. Future work will also extend the proposed framework beyond hourly bolus insulin and carbohydrate recommendations. The current action space captures important behavioral components, but it does not explicitly model other lifestyle factors such as sleep patterns, physical activity, or stress. Incorporating these behaviors as multi-scale lifestyle actions would require designing a separate action space operating at longer temporal resolutions (e.g., daily). While real-world clinical validation is essential, conducting such studies requires substantial time, regulatory approval (e.g., FDA clearance), patient recruitment, and financial resources. Therefore, near-term work will focus on developing a simulator that more accurately reflects patient behavior, adherence, and real-world variability, thereby providing a practical intermediate step toward translational evaluation. Building on this foundation, longer-term work will integrate GUIDE with AID technologies to evaluate its interaction with these systems and advance toward prospective studies involving individuals with T1D.

\section{Conclusion}
\label{sec:conclusion}

In this work, we proposed GUIDE, an RL-based framework that provides personalized behavioral decision support for T1D management. Unlike existing RL frameworks that primarily focus on insulin-only treatment, GUIDE extends the action space to include carbohydrate intake and bolus insulin, together with the corresponding action magnitude and timing. To support policy training and evaluation, the simulation environment incorporates a patient-specific glucose predictor to model personalized glucose responses to the agent's actions. It also includes a bolus safeguard that constrains insulin-on-board, cumulative insulin exposure, bolus timing, and dose magnitude to reduce potentially unsafe bolus actions.

The evaluation included 25 individuals from the AZT1D dataset and 12 individuals from the OhioT1DM dataset. Among the evaluated algorithms, CQL-BC achieved the highest average time-in-range on both datasets. On AZT1D, CQL-BC achieved a TIR of \(84.18 \pm 19.89\%\) and a TBR of \(0.43 \pm 1.27\%\). On OhioT1DM, it achieved a TIR of \(77.64 \pm 8.81\%\) and a TBR of \(2.81 \pm 3.56\%\). Behavioral analysis yielded mean cosine similarities of \(0.767 \pm 0.128\) on AZT1D and \(0.774 \pm 0.162\) on OhioT1DM, indicating consistent structural alignment between CQL-BC actions and historical patient action patterns. For CQL-BC on AZT1D, the partial-adherence analysis further showed that TIR decreased from \(84.40 \pm 19.70\%\) under full adherence to \(78.19 \pm 22.68\%\) and \(68.45 \pm 24.37\%\) under \(75\%\) and \(50\%\) adherence, respectively.

These findings provide simulation-based evidence that conservative offline RL can support a structured behavioral action space extending beyond insulin-only control. However, all reported outcomes were obtained within personalized learned simulation environments and do not establish real-world clinical effectiveness or safety. GUIDE therefore provides a computational foundation for investigating integrated behavioral decision support, while prospective clinical evaluation remain necessary before real-world use.

\section{Acknowledgment}
\label{sec:acknowledgment}

This work was supported in part by the National Science Foundation (NSF) under Award IIS-2402650. A portion of this work has taken place in the Learning Agents Research Group (LARG) and the Rewarding Lab at UT Austin. LARG research is supported in part by NSF (FAIN-2019844, NRT-2125858, OIC-2535195), ONR (N00014-24-1-2550), ARO (W911NF-17-2-0181, W911NF-23-2-0004, W911NF-25-1-0065), Lockheed Martin, Lyda Hill, and Good Systems. During this project, the Rewarding Lab has been supported by NSF (IIS-2402650, 2535195), ARO (W911NF-25-1-0254), Emerson, EA Ventures, UT Austin's Good Systems grand challenge, and Coefficient Giving. Peter Stone serves as the Chief Scientist of Sony CTC and receives financial compensation for that role. The terms of this arrangement have been reviewed and approved by the University of Texas at Austin in accordance with its policy on objectivity in research. Any opinions, findings, conclusions, or recommendations expressed in this material are those of the authors and do not necessarily reflect the views of the funding organizations.

\section*{References}
\bibliographystyle{IEEEtran}
\bibliography{ref}

\begin{thebibliography}{10}
\providecommand{\url}[1]{#1}
\csname url@samestyle\endcsname
\providecommand{\newblock}{\relax}
\providecommand{\bibinfo}[2]{#2}
\providecommand{\BIBentrySTDinterwordspacing}{\spaceskip=0pt\relax}
\providecommand{\BIBentryALTinterwordstretchfactor}{4}
\providecommand{\BIBentryALTinterwordspacing}{\spaceskip=\fontdimen2\font plus
\BIBentryALTinterwordstretchfactor\fontdimen3\font minus \fontdimen4\font\relax}
\providecommand{\BIBforeignlanguage}[2]{{%
\expandafter\ifx\csname l@#1\endcsname\relax
\typeout{** WARNING: IEEEtran.bst: No hyphenation pattern has been}%
\typeout{** loaded for the language `#1'. Using the pattern for}%
\typeout{** the default language instead.}%
\else
\language=\csname l@#1\endcsname
\fi
#2}}
\providecommand{\BIBdecl}{\relax}
\BIBdecl

\bibitem{popoviciu2023type}
M.~S. Popoviciu, N.~Kaka, Y.~Sethi, N.~Patel, H.~Chopra, and S.~Cavalu, ``Type 1 diabetes mellitus and autoimmune diseases: a critical review of the association and the application of personalized medicine,'' \emph{Journal of Personalized Medicine}, vol.~13, no.~3, p. 422, 2023.

\bibitem{o2024shifting}
A.~J. O’Donovan, S.~Gorelik, and L.~M. Nally, ``Shifting the paradigm of type 1 diabetes: a narrative review of disease-modifying therapies,'' \emph{Frontiers in Endocrinology}, vol.~15, p. 1477101, 2024.

\bibitem{ogle2025global}
G.~D. Ogle, F.~Wang, A.~Haynes, G.~A. Gregory, T.~W. King, K.~Deng, D.~Dabelea, S.~James, A.~J. Jenkins, X.~Li \emph{et~al.}, ``Global type 1 diabetes prevalence, incidence, and mortality estimates 2025: Results from the international diabetes federation atlas, and the t1d index version 3.0,'' \emph{Diabetes Research and Clinical Practice}, p. 112277, 2025.

\bibitem{battelino2019clinical}
T.~Battelino, T.~Danne, R.~M. Bergenstal, S.~A. Amiel, R.~Beck, T.~Biester, E.~Bosi, B.~A. Buckingham, W.~T. Cefalu, K.~L. Close \emph{et~al.}, ``Clinical targets for continuous glucose monitoring data interpretation: recommendations from the international consensus on time in range,'' \emph{Diabetes Care}, vol.~42, no.~8, pp. 1593--1603, 2019.

\bibitem{foster2019state}
N.~C. Foster, R.~W. Beck, K.~M. Miller, M.~A. Clements, M.~R. Rickels, L.~A. DiMeglio, D.~M. Maahs, W.~V. Tamborlane, R.~Bergenstal, E.~Smith \emph{et~al.}, ``State of type 1 diabetes management and outcomes from the t1d exchange in 2016--2018,'' \emph{Diabetes Technology \& Therapeutics}, vol.~21, no.~2, pp. 66--72, 2019.

\bibitem{american20226}
A.~D. A. P.~P. Committee, ``6. glycemic targets: standards of medical care in diabetes—2022,'' \emph{Diabetes Care}, vol.~45, no. Supplement\_1, pp. S83--S96, 2022.

\bibitem{bekiari2018artificial}
E.~Bekiari, K.~Kitsios, H.~Thabit, M.~Tauschmann, E.~Athanasiadou, T.~Karagiannis, A.-B. Haidich, R.~Hovorka, and A.~Tsapas, ``Artificial pancreas treatment for outpatients with type 1 diabetes: systematic review and meta-analysis,'' \emph{BMJ}, vol. 361, 2018, article e2094.

\bibitem{templer2022closed}
S.~Templer, ``Closed-loop insulin delivery systems: past, present, and future directions,'' \emph{Frontiers in Endocrinology}, vol.~13, p. 919942, 2022.

\bibitem{rodbard2016continuous}
D.~Rodbard, ``Continuous glucose monitoring: a review of successes, challenges, and opportunities,'' \emph{Diabetes Technology \& Therapeutics}, vol.~18, no. 2\_suppl, pp. S2--3, 2016.

\bibitem{limbert2024automated}
C.~Limbert, A.~J. Kowalski, and T.~P. Danne, ``Automated insulin delivery: a milestone on the road to insulin independence in type 1 diabetes,'' \emph{Diabetes Care}, vol.~47, no.~6, pp. 918--920, 2024.

\bibitem{steil2013algorithms}
G.~M. Steil, ``Algorithms for a closed-loop artificial pancreas: the case for proportional-integral-derivative control,'' \emph{Journal of Diabetes Science and Technology}, vol.~7, no.~6, pp. 1621--1631, 2013.

\bibitem{bequette2013algorithms}
B.~W. Bequette, ``Algorithms for a closed-loop artificial pancreas: the case for model predictive control,'' \emph{Journal of Diabetes Science and Technology}, vol.~7, no.~6, pp. 1632--1643, 2013.

\bibitem{bombaci2025impact}
B.~Bombaci, M.~Calderone, A.~Di~Pisa, M.~La~Rocca, A.~Torre, F.~Lombardo, G.~Salzano, and S.~Passanisi, ``Impact of automated insulin delivery systems in children and adolescents with type 1 diabetes previously treated with multiple daily injections: a single-center real-world study,'' \emph{Medicina}, vol.~61, no.~9, p. 1602, 2025.

\bibitem{tejedor2020reinforcement}
M.~Tejedor, A.~Z. Woldaregay, and F.~Godtliebsen, ``Reinforcement learning application in diabetes blood glucose control: A systematic review,'' \emph{Artificial Intelligence in Medicine}, vol. 104, p. 101836, 2020.

\bibitem{denes2024reinforcement}
L.~D{\'e}nes-Fazakas, L.~Szil{\'a}gyi, L.~Kov{\'a}cs, A.~De~Gaetano, and G.~Eigner, ``Reinforcement learning: a paradigm shift in personalized blood glucose management for diabetes,'' \emph{Biomedicines}, vol.~12, no.~9, p. 2143, 2024.

\bibitem{bothe2013use}
M.~K. Bothe, L.~Dickens, K.~Reichel, A.~Tellmann, B.~Ellger, M.~Westphal, and A.~A. Faisal, ``The use of reinforcement learning algorithms to meet the challenges of an artificial pancreas,'' \emph{Expert Review of Medical Devices}, vol.~10, no.~5, pp. 661--673, 2013.

\bibitem{aria2025learning}
H.~P. Aria, H.~Kim, S.~M. Alsadat, and Z.~Xu, ``Learning causal dynamics and reward machines: A framework for faster reinforcement learning with extended temporal tasks,'' in \emph{2025 5th International Conference on Computer, Control and Robotics (ICCCR)}.\hskip 1em plus 0.5em minus 0.4em\relax IEEE, 2025, pp. 519--525.

\bibitem{fox2020deep}
I.~Fox, J.~Lee, R.~Pop-Busui, and J.~Wiens, ``Deep reinforcement learning for closed-loop blood glucose control,'' in \emph{Machine Learning for Healthcare Conference}.\hskip 1em plus 0.5em minus 0.4em\relax PMLR, 2020, pp. 508--536.

\bibitem{hettiarachchi2024g2p2c}
C.~Hettiarachchi, N.~Malagutti, C.~J. Nolan, H.~Suominen, and E.~Daskalaki, ``G2p2c—a modular reinforcement learning algorithm for glucose control by glucose prediction and planning in type 1 diabetes,'' \emph{Biomedical Signal Processing and Control}, vol.~90, p. 105839, 2024.

\bibitem{lee2020toward}
S.~Lee, J.~Kim, S.~W. Park, S.-M. Jin, and S.-M. Park, ``Toward a fully automated artificial pancreas system using a bioinspired reinforcement learning design: In silico validation,'' \emph{IEEE Journal of Biomedical and Health Informatics}, vol.~25, no.~2, pp. 536--546, 2020.

\bibitem{shah2016insulin}
R.~B. Shah, M.~Patel, D.~M. Maahs, and V.~N. Shah, ``Insulin delivery methods: Past, present and future,'' \emph{International Journal of Pharmaceutical Investigation}, vol.~6, no.~1, p.~1, 2016.

\bibitem{doyle2014closed}
F.~J. Doyle~III, L.~M. Huyett, J.~B. Lee, H.~C. Zisser, and E.~Dassau, ``Closed-loop artificial pancreas systems: engineering the algorithms,'' \emph{Diabetes Care}, vol.~37, no.~5, pp. 1191--1197, 2014.

\bibitem{ramkissoon2018unannounced}
C.~M. Ramkissoon, P.~Herrero, J.~Bondia, and J.~Vehi, ``Unannounced meals in the artificial pancreas: detection using continuous glucose monitoring,'' \emph{Sensors}, vol.~18, no.~3, p. 884, 2018.

\bibitem{emerson2023offline}
H.~Emerson, M.~Guy, and R.~McConville, ``Offline reinforcement learning for safer blood glucose control in people with type 1 diabetes,'' \emph{Journal of Biomedical Informatics}, vol. 142, p. 104376, 2023.

\bibitem{fujimoto2018addressing}
S.~Fujimoto, H.~Hoof, and D.~Meger, ``Addressing function approximation error in actor-critic methods,'' in \emph{International Conference on Machine Learning}.\hskip 1em plus 0.5em minus 0.4em\relax PMLR, 2018, pp. 1587--1596.

\bibitem{fujimoto2019off}
S.~Fujimoto, D.~Meger, and D.~Precup, ``Off-policy deep reinforcement learning without exploration,'' in \emph{International Conference on Machine Learning}.\hskip 1em plus 0.5em minus 0.4em\relax PMLR, 2019, pp. 2052--2062.

\bibitem{kumar2020conservative}
A.~Kumar, A.~Zhou, G.~Tucker, and S.~Levine, ``Conservative q-learning for offline reinforcement learning,'' \emph{Advances in Neural Information Processing Systems}, vol.~33, pp. 1179--1191, 2020.

\bibitem{adjevi2026explainable}
A.~Adjevi, A.~M. Abdirashid, F.~Akta{\c{s}}, M.~H.~B. Ucar, and S.~Solak, ``Explainable reinforcement learning for glucose monitoring based on shapley value analysis,'' \emph{Computer Methods and Programs in Biomedicine}, p. 109266, 2026.

\bibitem{marchetti2025deep}
A.~Marchetti, D.~Sasso, F.~D’Antoni, F.~Morandin, M.~Parton, M.~A.~G. Matarrese, and M.~Merone, ``Deep reinforcement learning for type 1 diabetes: Dual ppo controller for personalized insulin management,'' \emph{Computers in Biology and Medicine}, vol. 191, p. 110147, 2025.

\bibitem{man2014uva}
C.~D. Man, F.~Micheletto, D.~Lv, M.~Breton, B.~Kovatchev, and C.~Cobelli, ``The uva/padova type 1 diabetes simulator: new features,'' \emph{Journal of Diabetes Science and Technology}, vol.~8, no.~1, pp. 26--34, 2014.

\bibitem{khamesian2026glycemic}
S.~Khamesian, A.~Arefeen, M.~A. Grando, B.~M. Thompson, and H.~Ghasemzadeh, ``Glycemic-aware and architecture-agnostic training framework for blood glucose forecasting in type 1 diabetes,'' \emph{BMC Artificial Intelligence}, vol.~2, no.~1, p.~13, 2026.

\bibitem{yamagata2020model}
T.~Yamagata, A.~O'Kane, A.~Ayobi, D.~Katz, K.~Stawarz, P.~Marshall, P.~Flach, and R.~Santos-Rodr{\'\i}guez, ``Model-based reinforcement learning for type 1 diabetes blood glucose control,'' \emph{arXiv preprint arXiv:2010.06266}, 2020.

\bibitem{alsadat2025trajectory}
S.~M. Alsadat and Z.~Xu, ``Trajectory-based automata learning for offline reinforcement learning,'' in \emph{2025 American Control Conference (ACC)}.\hskip 1em plus 0.5em minus 0.4em\relax IEEE, 2025, pp. 3473--3478.

\bibitem{schulman2017proximal}
J.~Schulman, F.~Wolski, P.~Dhariwal, A.~Radford, and O.~Klimov, ``Proximal policy optimization algorithms,'' \emph{arXiv preprint arXiv:1707.06347}, 2017.

\bibitem{haarnoja2018soft}
T.~Haarnoja, A.~Zhou, P.~Abbeel, and S.~Levine, ``Soft actor-critic: Off-policy maximum entropy deep reinforcement learning with a stochastic actor,'' in \emph{International Conference on Machine Learning}.\hskip 1em plus 0.5em minus 0.4em\relax PMLR, 2018, pp. 1861--1870.

\bibitem{khamesian2025azt1d}
S.~Khamesian, A.~Arefeen, B.~M. Thompson, M.~A. Grando, and H.~Ghasemzadeh, ``Azt1d: A real-world dataset for type 1 diabetes,'' in \emph{2025 IEEE 21st International Conference on Body Sensor Networks (BSN)}.\hskip 1em plus 0.5em minus 0.4em\relax IEEE, 2025, pp. 1--4.

\bibitem{guide_supplement}
\BIBentryALTinterwordspacing
S.~Khamesian \emph{et~al.} (2026) Supplementary material for ``{GUIDE}: Reinforcement learning for behavioral action support in type 1 diabetes''. GitHub. [Online]. Available: \url{https://github.com/SamanKhamesian/GUIDE/blob/main/material/supplementary_material.pdf}
\BIBentrySTDinterwordspacing

\bibitem{marling2020ohiot1dm}
C.~Marling and R.~Bunescu, ``The ohiot1dm dataset for blood glucose level prediction: Update 2020,'' in \emph{CEUR Workshop Proceedings}, vol. 2675, 2020, p.~71.

\bibitem{lim2021blood}
M.~H. Lim, W.~H. Lee, B.~Jeon, and S.~Kim, ``A blood glucose control framework based on reinforcement learning with safety and interpretability: In silico validation,'' \emph{IEEE Access}, vol.~9, pp. 105\,756--105\,775, 2021.

\bibitem{danne2017international}
T.~Danne, R.~Nimri, T.~Battelino, R.~M. Bergenstal, K.~L. Close, J.~H. DeVries, S.~Garg, L.~Heinemann, I.~Hirsch, S.~A. Amiel \emph{et~al.}, ``International consensus on use of continuous glucose monitoring,'' \emph{Diabetes Care}, vol.~40, no.~12, pp. 1631--1640, 2017.

\bibitem{preisser2011multiple}
J.~S. Preisser, P.~K. Sen, and S.~Offenbacher, ``Multiple hypothesis testing for experimental gingivitis based on wilcoxon signed rank statistics,'' \emph{Statistics in Biopharmaceutical Research}, vol.~3, no.~2, pp. 372--384, 2011.

\bibitem{lee2018proper}
S.~Lee and D.~K. Lee, ``What is the proper way to apply the multiple comparison test?'' \emph{Korean Journal of Anesthesiology}, vol.~71, no.~5, pp. 353--360, 2018.

\end{thebibliography}

\end{document}